\documentclass{article}

\usepackage[preprint, nonatbib]{neurips_2021}

\usepackage[utf8]{inputenc} 
\usepackage[T1]{fontenc}    
\usepackage{url}            
\usepackage{booktabs}       
\usepackage{amsfonts}       
\usepackage{nicefrac}       
\usepackage{microtype}      
\usepackage{xcolor}         

\usepackage{float}
\usepackage{graphicx}
\usepackage{multirow}
\usepackage{framed}
\usepackage{amsmath}
\usepackage{amssymb}
\usepackage{xcolor}
\usepackage{tikz}
\usepackage{subfig}
\usepackage{bm}
\usepackage{enumitem}
\usepackage{wrapfig}
\usepackage{algorithmic}
\usepackage{textcomp}
\usepackage{booktabs} 
\usepackage{tikz}
\usepackage{colortbl}
\usepackage{listings}

\usepackage[pagebackref=true]{hyperref} 
\renewcommand*\backref[1]{\ifx#1\relax \else (Cited on page #1) \fi}

\definecolor{codegreen}{rgb}{0,0.6,0}
\definecolor{codegray}{rgb}{0.5,0.5,0.5}
\definecolor{codepurple}{rgb}{0.58,0,0.82}
\definecolor{backcolour}{rgb}{0.95,0.95,0.92}

\lstdefinestyle{mystyle}{
  backgroundcolor=\color{backcolour}, commentstyle=\color{codegreen},
  keywordstyle=\color{magenta},
  numberstyle=\tiny\color{codegray},
  stringstyle=\color{codepurple},
  basicstyle=\ttfamily\footnotesize,
  breakatwhitespace=false,         
  breaklines=true,                 
  captionpos=b,                    
  keepspaces=true,                 
  numbers=left,                    
  numbersep=5pt,                  
  showspaces=false,                
  showstringspaces=false,
  showtabs=false,                  
  tabsize=2
}

\newcommand{\red}[1]{\textcolor{red}{#1}}
\newcommand{\blue}[1]{\textcolor{blue}{#1}}

\newcommand{\green}[1]{\textcolor{green}{#1}}

\title{On Representation Knowledge Distillation for \\Graph Neural Networks}

\author{%
  Chaitanya K. Joshi\thanks{CKJ is now at University of Cambridge, UK. Email: \texttt{chaitanya.joshi@cl.cam.ac.uk}} , Fayao Liu, Xu Xun, Jie Lin, Chuan Sheng Foo \\
  Institute for Infocomm Research, A*STAR, Singapore
}

\begin{document}

\maketitle

\begin{abstract}

Knowledge distillation is a learning paradigm for boosting resource-efficient graph neural networks (GNNs) using more expressive yet cumbersome teacher models.
Past work on distillation for GNNs proposed the Local Structure Preserving loss (LSP), which matches \textit{local} structural relationships defined over edges across the student and teacher's node embeddings.
This paper studies whether preserving the \textit{global} topology of how the teacher embeds graph data can be a more effective distillation objective for GNNs, as real-world graphs often contain latent interactions and noisy edges. 
We propose Graph Contrastive Representation Distillation (G-CRD), which uses contrastive learning to \textit{implicitly} preserve global topology by aligning the student node embeddings to those of the teacher in a shared representation space.
Additionally, we introduce an expanded set of benchmarks on large-scale real-world datasets where the performance gap between teacher and student GNNs is non-negligible.
Experiments across 4 datasets and 14 heterogeneous GNN architectures show that G-CRD consistently boosts the performance and robustness of lightweight GNNs, outperforming LSP (and a global structure preserving variant of LSP) as well as baselines from 2D computer vision.
An analysis of the representational similarity among teacher and student embedding spaces reveals that G-CRD balances preserving local and global relationships, while structure preserving approaches are best at preserving one or the other.
Our code is available at \url{https://github.com/chaitjo/efficient-gnns}

\end{abstract}


\section{Introduction}
\label{sec:intro}

Graph Neural Networks (GNNs) \cite{kipf2017semi, hamilton2017inductive, wang2019dynamic, velickovic2018graph, battaglia2018relational} generalize convolutional networks from 2D computer vision to irregular data structures such as graphs, sets, and 3D point clouds. 
Recent years have seen impactful applications of GNNs in fields ranging from social networks \cite{ying2018graph, monti2019fake} to biomedicine \cite{stokes2020deep, gainza2020deciphering, long2020predicting}.
While the community has recently focused on large-scale data \cite{hu2020open, hu2021ogb, addanki2021large}, more expressive architectures \cite{xu2018how, li2019deepgcns, corso2020principal, dwivedi2020benchmarking} as well as improving generalization via self-supervised learning \cite{hu2019strategies, you2020graph, qiu2020gcc, jiao2020subgraph},
there has been an emerging line of work on lightweight, resource-efficient GNNs \cite{sign_icml_grl2020, tailor2021adaptive, li2021gnn1000, gao2020graph, zhao2021probabilistic, zhao2020learned, tailor2021degreequant, Zeng2020GraphSAINT:} that achieve high performance under computation and memory constraints.
Boosting the performance and reliability of lightweight models is critical for accelerating a myriad of applications, including real-time recommendations on social networks,
safety-critical 3D perception for autonomous robots,
and encrypted models for proprietary biomedical data. 

\begin{figure*}[t!]
 \subfloat[Representation distillation pipeline (teacher and student features are projected onto planes for visualization)]{
	\begin{minipage}[c]{\textwidth}
	   \centering
	   \includegraphics[width=0.7\textwidth]{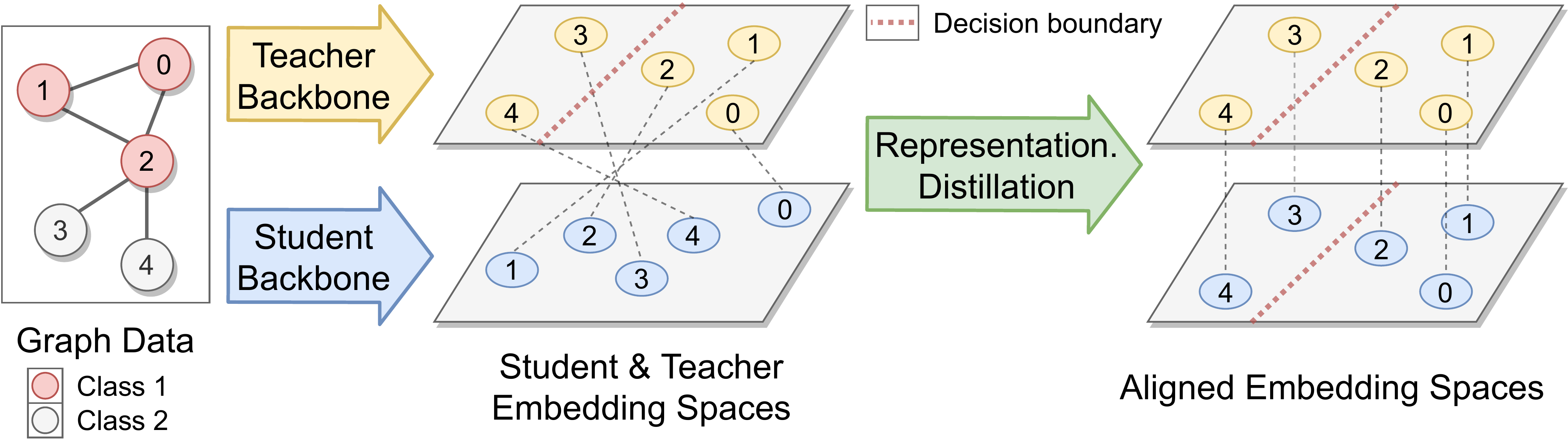}
	   \end{minipage}
	}
 \newline
 \subfloat[LSP]{
	\begin{minipage}[c][]{0.3\textwidth}
	   \centering
	   \includegraphics[width=1\textwidth]{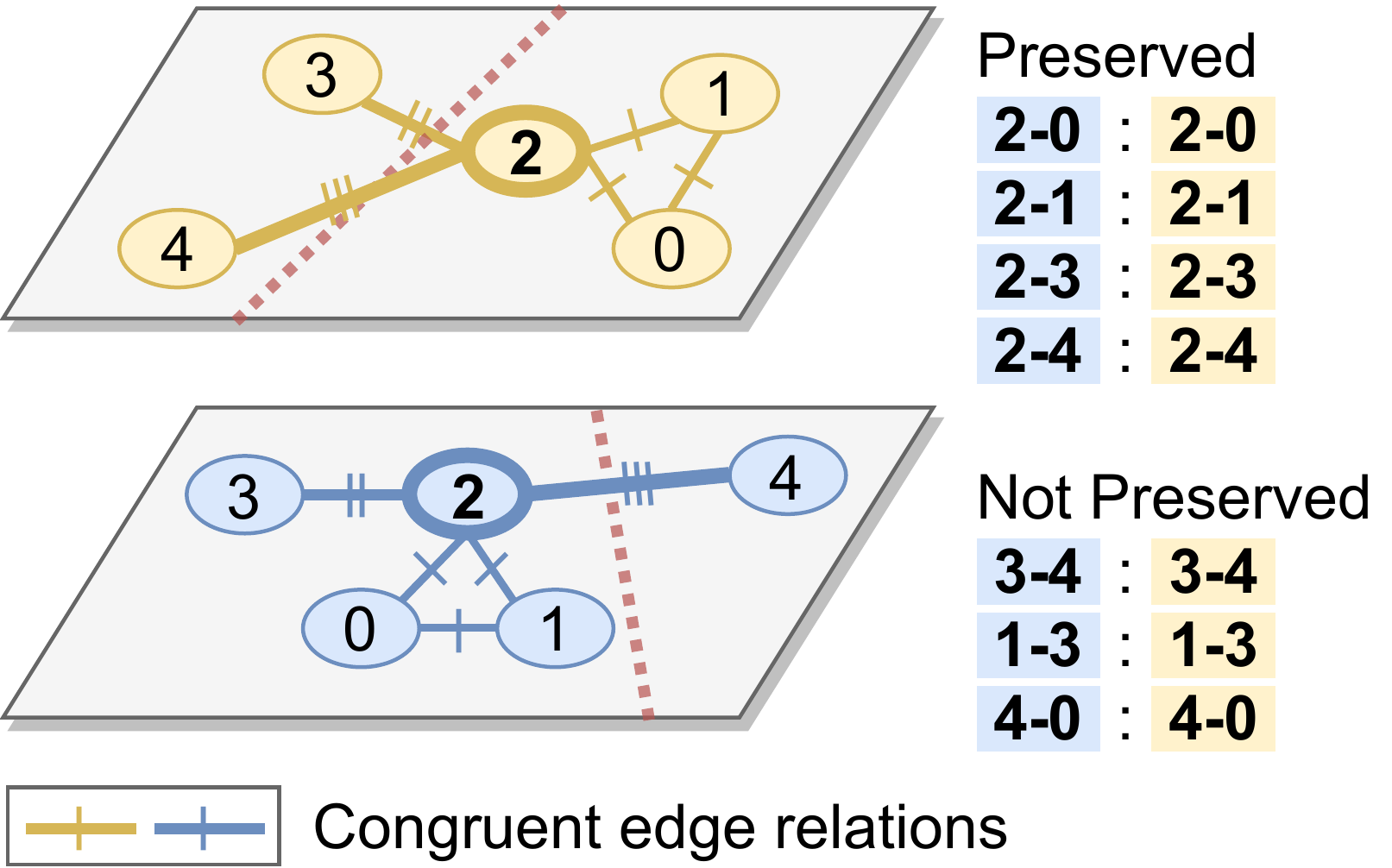}
	\end{minipage}
	}
 \hfill
 \subfloat[GSP]{
	\begin{minipage}[c][]{0.3\textwidth}
	   \centering
	   \includegraphics[width=1\textwidth]{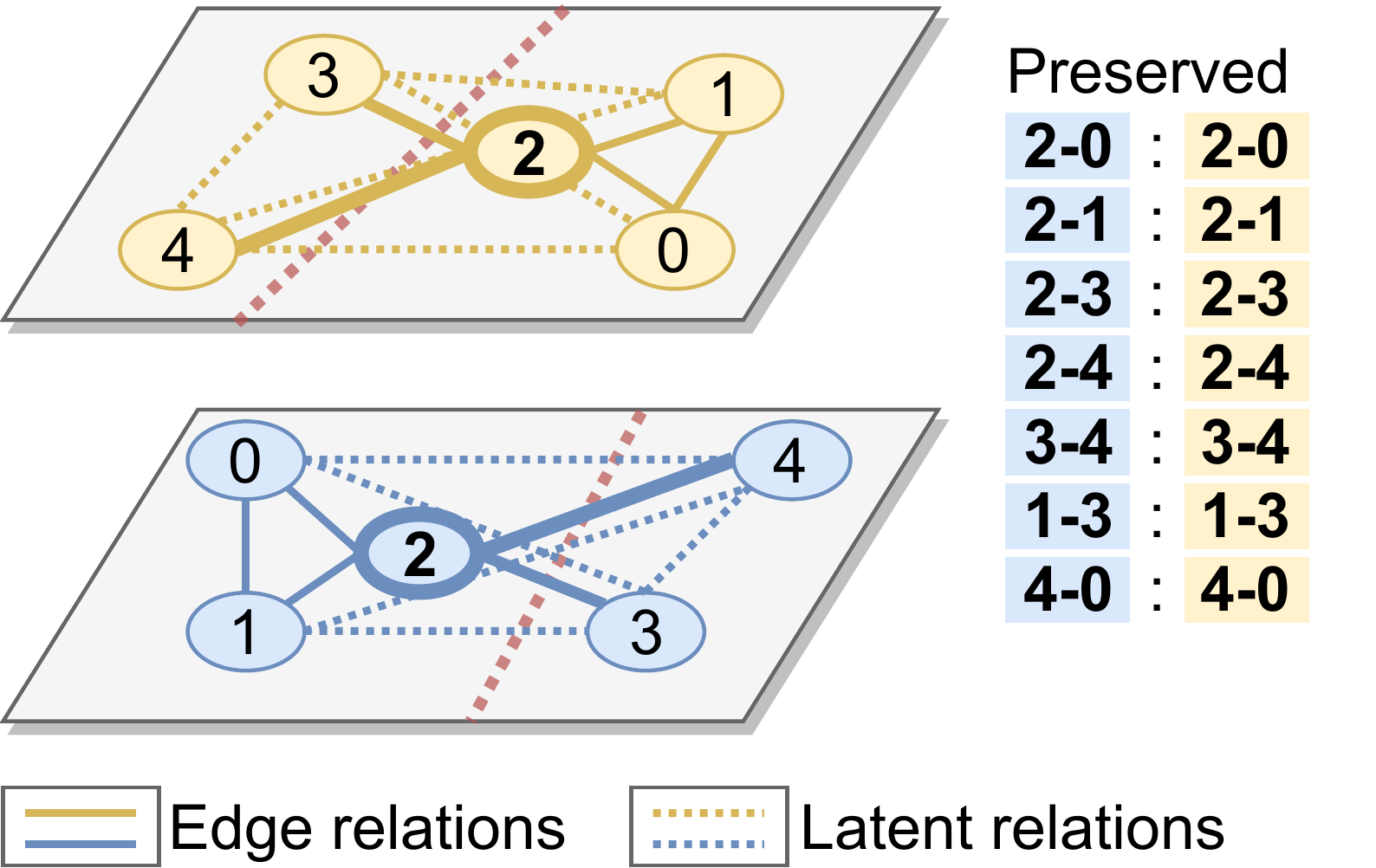}
	\end{minipage}
	}
 \hfill
 \subfloat[G-CRD]{
	\begin{minipage}[c][]{0.3\textwidth}
	   \centering
	   \includegraphics[width=1\textwidth]{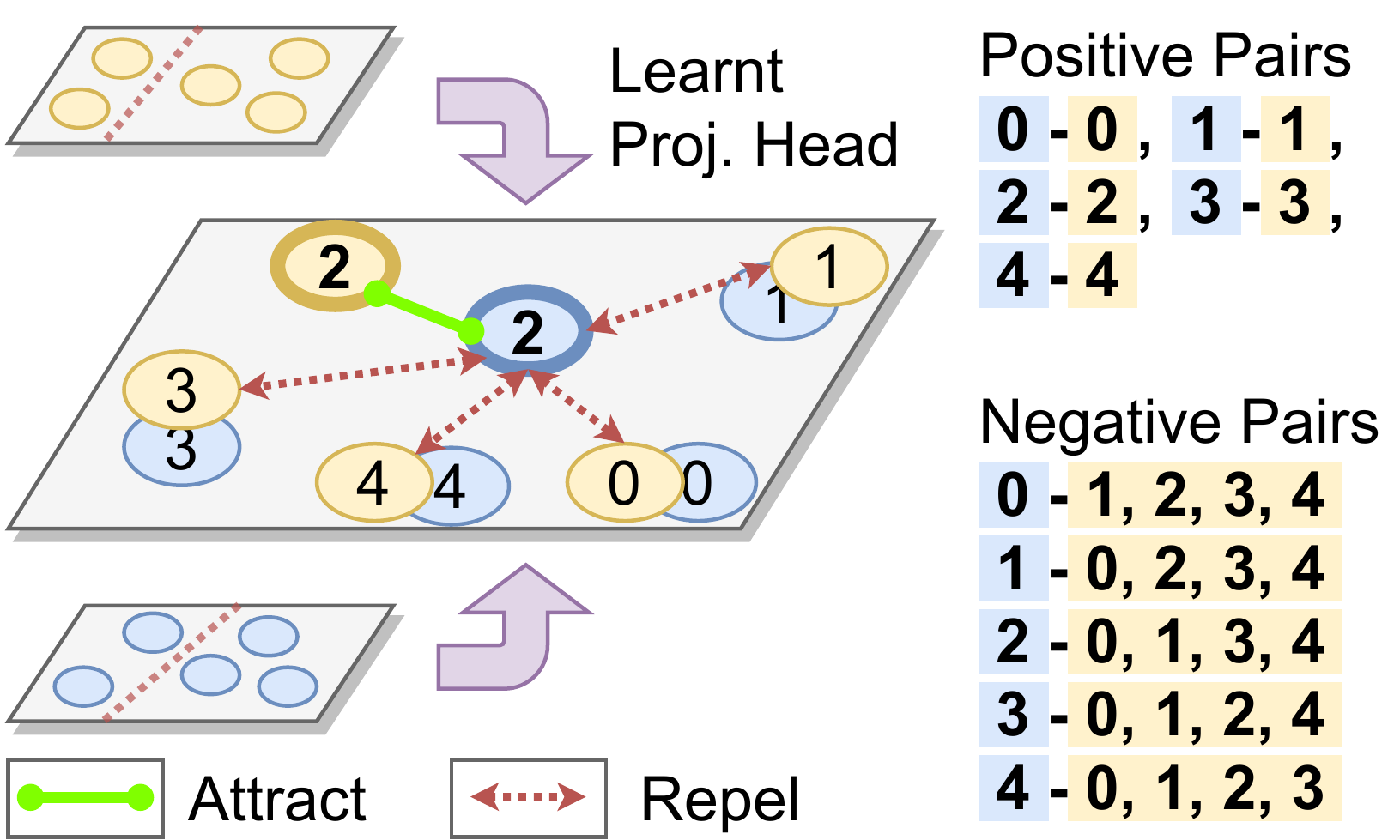}
	\end{minipage}
	}
\caption{
\textbf{Graphical overview of representation distillation for GNNs.}
(a) Representational knowledge is transferred by aligning the latent node embedding space of the student to that of the teacher.
(b) \textbf{Local Structure Preserving} loss \cite{yang2020distilling} considers pairwise relationships over graph edges, but may not preserve global topology of the teacher's embedding space due to latent interactions among disconnected nodes.
(c) Preserving all pairwise relationships, \textit{i.e.} \textbf{Global Structure Preserving} loss, better preserves the global topology of the embedding space, but can be challenging to optimize/scale up due to an explosion of possible pairs.
(d) Different from relation-based approaches, \textbf{Graph Contrastive Representation Distillation} transfers representational knowledge by contrastive learning among positive/negative pairwise relations \textit{across} the teacher and student embedding spaces via learnt projection heads.
}
\label{fig:main}
\end{figure*}

A promising and generic approach for improving lightweight deep learning models is Knowledge Distillation (KD) \cite{hinton2015distilling}. 
KD is a teacher-student learning paradigm that transfers knowledge from high performance but resource-intensive teacher models to resource-efficient students.
Pioneered by Hinton et al. \cite{hinton2015distilling}, \textit{logit}-based distillation trains the student to match the output logits of teachers, in addition to standard supervised learning. 
Recent work has attempted to go beyond logit-based distillation by transferring \textit{representational} knowledge from the teacher to the student through the design of loss functions that align the latent embedding spaces of the teacher and student \cite{gou2021knowledge}, see Fig.\ref{fig:main}(a) for an intuitive overview. 


Unlike for 2D image data, representation distillation for GNNs has largely been unexplored besides the pioneering work of Yang et al. \cite{yang2020distilling} that proposes the \textbf{Local Structure Preserving} (LSP) objective.
LSP encourages the student to mimic the pairwise similarities with immediate neighbours present in the teacher's node embedding space. 
Thus, structural similarities are preserved over \textit{local} graph edges.
However, this is not guaranteed to preserve the \textit{global} topology of how the teacher embeds graphs as this ignores latent interactions among disconnected nodes, see Fig.\ref{fig:main}(b). 

Modelling latent interactions is often critical for solving tasks on real-world graphs, which tend to be incomplete or noisy as edges are determined heuristically.
Expressive and deep GNNs are able to better capture the full structure of the underlying data, including latent interactions, in their node embedding space.
Learning the same information via shallow and efficient GNNs may not be possible due to be mismatch in representational capacity.
Thus, the ideal representation distillation technique would transfer \textit{global} structural information from expressive teachers to lightweight students.

To preserve global relationships, it is natural to extend LSP to \textit{explicitly} consider all possible pairwise similarities among node features.
However, we found this \textbf{Global Structure Preserving} (GSP) approach to be challenging to scale as the number of possible relationships may explode, see Fig.\ref{fig:main}(c). 


Thus, we introduce a new objective that \textit{implicitly} preserves global topology by aligning the student and teacher node feature vectors via contrastive learning \cite{oord2018representation, wu2018unsupervised, chen2020simple}. 
We formulate our objective as a node-level contrastive task on pairwise relationships \textit{across} the teacher and student embedding spaces, see Fig.\ref{fig:main}(d). 
We term this objective \textbf{Graph Contrastive Representation Distillation (G-CRD)}, as it generalizes CRD \cite{tian2019contrastive} from sample-level 2D image classification tasks to fine-grained node-level tasks on graphs.
G-CRD preserves global relationships by training the student to spatially align its node embeddings with the corresponding teacher node embedding, termed positive samples. 
For \textit{e.g.}, in Fig.\ref{fig:main}(d), the student's embedding for node 2 is pushed to the teacher's embedding for node 2 by maximizing the learnt similarity metric.
Additionally, the student's embedding for node 2 is repelled from all other node embeddings from the teacher, termed negative samples, by minimizing their similarities.
The number of negative samples can be varied to ensure scalability.


We compare G-CRD to LSP, its global variant GSP, and baseline techniques from 2D computer vision across a diverse range of model architectures and tasks from Open Graph Benchmarks \cite{hu2020open} and S3DIS \cite{armeni20163d}.
Crucially, our evaluation focuses on out-of-distribution generalization on real-world data, where the performance gap between cumbersome teachers and lightweight students is non-negligible. 
We believe this is critical for testing the efficacy and robustness of knowledge distillation, but was missing from the LSP study \cite{yang2020distilling} which used small-scale and synthetic datasets with negligible performance gaps between teachers and students.
In addition to our technical contributions, we hope that our expanded set of benchmarks can form the basis for comparison in future work.

\textbf{Our contributions are summarized as follows:}
\begin{itemize}[itemsep=0em]
    \item We study GNN representation distillation from the perspective of preserving global topology. 
    We introduce Graph Contrastive Representation Distillation (G-CRD), the first contrastive distillation technique specialized for GNNs which trains students to implicitly preserve the global topology of the teacher's node embedding space.
    \item We benchmark GNN distillation on large-scale datasets which test out-of-distribution generalization.
    Our experiments compare 6 distillation techniques across 4 tasks and 14 architectures for teachers and students.
    \item Training lightweight GNNs with G-CRD consistently outperform the structure preserving objectives, LSP and GSP, as well as baselines adapted from 2D computer vision.
    We further analyze the robustness, transferability, quantizability, and representational similarity of teacher and student embeddings in order to unpack the efficacy of G-CRD.
\end{itemize}


\section{Preliminaries}
\label{sec:method}

\subsection{Graph Representation Learning}

Graph Neural Networks (GNNs) take as input an unordered set of nodes and the graph connectivity among them, and learn latent node representations for them via iterative feature aggregation or message passing across local neighborhoods \cite{battaglia2018relational}.
Consider a graph $\mathcal{G} = \left( \mathcal{V}, \mathcal{E} \right)$, where $\mathcal{V}$ is a set of $n$ nodes, and $\mathcal{E}$ is a set of edges associated with the nodes.
Each node $i \in \mathcal{V}$ is associated with a $d$-dimensional initial feature vector $f_i$, which is iteratively updated by the backbone GNN via a generic message passing or graph convolution operation:
\begin{equation}
    \label{eq:mpnn}
    f_i^{\ell+1} = \textsc{Upd} \Bigg( f_i^{\ell}, \underset{(i,j) \in \mathcal{E}}{\textsc{Agg}} \Big( \textsc{Msg} \big( f_i^{\ell}, f_j^{\ell}, e_{ij} \big) \Big) \Bigg),
\end{equation}
where $\textsc{Msg}, \textsc{Upd}$ are learnable transformations such as multi-layer perceptrons, $e_{ij}$ are optional edge features, and $\textsc{Agg}$ is a permutation-invariant local neighborhood aggregation function such as summation, maximization, averaging, or weighted averaging via attention.

We are usually interested in making predictions for each node in the graph, \textit{e.g.} node classification on social networks or semantic segmentation of 3D point clouds.
Thus, the final feature vectors after $L$ layers of message passing $f_i^{L}$ are passed to a linear classifier to obtain logits $z_i$, and the neural network is trained end-to-end via a cross-entropy loss $\mathcal{H}$ with the groundtruth class labels $y_i$: $ \mathcal{L_{\textsc{Sup}}} = \mathcal{H} \left( y_i, \text{softmax} \left( z_i \right) \right)$.
For graph-level tasks, such as molecular property prediction, the feature vectors of all nodes are pooled via a permutation-invariant function $\textsc{Pool}$ to obtain a global feature vector $f_G = \textsc{Pool} \left( f_i^{L} | i \in \mathcal{V} \right)$, and then passed to a linear classifier. 
Notably, representation learning still occurs at the node-level for graph-level prediction tasks \cite{mesquita2020rethinking}, unlike 2D image classification where pooling layers learn feature vectors for entire images.

\textbf{GNNs and 3D point clouds.}
The message passing framework in \eqref{eq:mpnn} can be used to present a unified `geometric' view of deep learning on non-Euclidean data structures such as 3D point clouds, irregular voxel grids, meshes, and sets \cite{bronstein2021geometric, joshi2020transformers}. 
In this work, we consider 3D point cloud networks \cite{qi2017pointnet++, thomas2019KPConv} which process sets of points as nodes.
The edges originating from each point are heuristically determined by the $k$-nearest neighbors or radius ball queries in the 3D coordinate space.
We additionally consider sparse 3D voxel networks \cite{choy20194d, liu2019point, tang2020searching} and use trilinear interpolation to convert voxel-level features to point-level features before distillation.


\subsection{Knowledge Distillation}

Knowledge Distillation (KD) transfers \textit{dark knowledge} from high capacity teacher models (which are cumbersome to deploy) to efficient students via matching their output logits in addition to supervised learning.
At each node $i \in \mathcal{V}$, logit-based KD \cite{hinton2015distilling} uses the cross-entropy loss or KL-divergence to match the output logits of the student $z_i^{S}$ and teacher $z_i^{T}$, scaled by temperature $\tau_1$:
\begin{equation}
    \label{eq:kd}
    \mathcal{L}_{\text{KD}} = \sum_{i \in \mathcal{V}} \mathcal{H} \left( \text{softmax} \left( z_i^{T} / \tau_1 \right) , \text{softmax} \left( z_i^{S} / \tau_1 \right) \right).
\end{equation}


In this work, we study auxiliary representation distillation techniques that augment or replace logit-based distillation with \textit{representational knowledge} using the teacher and student features, $f^{T} \in \mathbb{R}^{d^T}$ and $f^{S} \in \mathbb{R}^{d^S}$, respectively.\footnote{
Following best practices \cite{tian2019contrastive,wang2020defense}, we consider feature vectors from the penultimate layer before the prediction head for representation distillation. 
}
Our overall objective function for training the student model via the logits as well as latent feature vectors of the teacher is as follows: 
\begin{equation}
    \label{eq:kd-sup-aux}
    \mathcal{L} = \left( 1 - \alpha \right) \mathcal{L}_{\textsc{Sup}} \ + \ \alpha \tau_1^2 \ \mathcal{L}_{\text{KD}} \ + \ \beta \ \mathcal{L}_{\text{Aux}},
\end{equation}
where $\alpha, \beta$ are balancing weights for the logit-based KD loss $\mathcal{L}_{\text{KD}}$ and auxiliary representation distillation loss $\mathcal{L}_{\text{Aux}}$, respectively.
Next, we will describe how $\mathcal{L}_{\text{Aux}}$ is instantiated as different representation distillation techniques such as  $\mathcal{L}_{\text{LSP}}$,  $\mathcal{L}_{\text{GSP}}$, and  $\mathcal{L}_{\text{G-CRD}}$.


\subsection{Local Structure Preserving Distillation}

We briefly summarize LSP \cite{yang2020distilling}, a GNN representation distillation objective which trains the student model to preserve the local structure of graph data from the teacher's node embedding space.
The local structure around each node is defined as the set of parameterized pairwise distances to its neighboring nodes in the latent feature space.
The similarity between a pair of linked nodes can be computed by kernel functions (which we tune for):
\begin{equation}
    \label{eq:lsp-sim}
    \mathcal{K} \left( f_i, f_j \right) =
    \begin{cases}
        \lVert f_i - f_j \rVert_{2}^{2}, & \text{Euclidean distance}, \\
        f_i' \cdot f_j, & \text{Linear kernel}, \\
        \left( f_i' \cdot f_j + c \right)^{d}, & \text{Polynomial kernel}, \\
        e^{ - \frac{1}{2\sigma} \lVert f_i - f_j \rVert^{2} }, & \text{RBF kernel}.
    \end{cases}
\end{equation}
Thus, the local structure around each node $i \in \mathcal{V}$ is the softmax probability distribution of the similarities among $f_i$ and its neighbors' features $f_j \ \forall (i,j) \in \mathcal{E}$.
LSP trains the student model to mimic the local structure from the teacher's embedding space via KL-divergence:
\begin{equation}
    \label{eq:lsp}
    \mathcal{L}_{\text{LSP}} = \sum_{i \in \mathcal{V}} \mathcal{D}_{\text{KL}} \Big( \underset{(i,j) \in \mathcal{E}}{\text{softmax}} \big( \mathcal{K} ( f_i^S, f_j^S ) \big) \, \Vert \, \underset{(i,j) \in \mathcal{E}}{\text{softmax}} \big( \mathcal{K} ( f_i^T, f_j^T ) \big) \Big).
\end{equation}


\subsection{Global Structure Preserving Distillation}

The purely local LSP objective over pre-defined edges does not preserve the global topology of how the teacher embeds the graph, as it does not account for latent interactions among disconnected nodes.
We would like to \textit{explicitly} train the student to preserve the global topology in order to better distill representational knowledge from the teacher.
To achieve this, we first introduce a simple extension of LSP, the Global Structure Preserving loss (GSP), which matches \textit{all} pairwise similarities among node features via mean squared error:
\begin{equation}
    \label{eq:gsp}
    \mathcal{L}_{\text{GSP}} = \sum_{i \in \mathcal{V}} \sum_{j \in \mathcal{V}} \big\lVert \mathcal{K} \left( f_i^T, f_j^T \right) - \mathcal{K} \left( f_i^S, f_j^S \right) \big\rVert^2_2.
\end{equation}
While theoretically more powerful than LSP, GSP may be computationally inefficient and involves two key design choices: (See App.\ref{app:gsp} for ablation studies.)

\textbf{MSE over KL-divergence} 
We omit normalizing the similarities via softmax followed by KL-divergence.
Instead, we use MSE to match the raw teacher and student pairwise similarity matrices as it worked better empirically.

\textbf{Scalability and sub-sampling.}
For a set of $n$ nodes, GSP needs to computer $n^2$ pairwise similarities for both the teacher and student features.
We often need to sub-sample large graphs or 3D point clouds when computing $\mathcal{L}_{\text{GSP}}$ due to GPU memory constraints instead of storing all possible pairwise similarities.
For each experiment, random sub-sampling was done to retain as many nodes as possible subject to GPU memory limitations.


\section{Graph Contrastive Representation Distillation}

An alternative to \textit{explicit} structure preserving techniques such as LSP and GSP is to directly align the node features from the student to those of the teacher.
An objective on the features would be a stronger constraint than over pairwise relationships, resulting in \textit{implicitly} preserving relationships over pre-defined as well as latent edges, and thereby preserving global topology.
However, we and others \cite{yang2020distilling} found direct feature mimicking approaches \cite{romero2014fitnets,zagoruyko2016paying} from 2D computer vision to be ineffective for GNNs due to mismatch in representational capacity.

Thus, we introduce Graph Contrastive Representation Distillation (G-CRD) which formulates representation distillation as a contrastive learning task on pairwise relationships \textit{across} the teacher and student embedding spaces.
Intuitively, we want to maximize the similarity among pairs of student and teacher feature vectors corresponding to the same node, \textit{i.e.} $f_i^{S} , f_i^T$ (positive samples), while pushing away the feature vectors of pairs of unmatched nodes, \textit{i.e.} $f_i^{S} , f_j^T$ (negative samples).
To achieve this, we adapt the InfoNCE objective \cite{oord2018representation,chen2020simple} to the teacher-student paradigm and pose representation distillation as the task of classifying positive pairs among the set of distractors via a temperature-scaled softmax function:
\begin{equation}
    \label{eq:G-CRD}
    \mathcal{L}_{\text{G-CRD}} = - \sum_{i \in \mathcal{V}} \text{ log} \frac{\text{ exp} \big( \frac{h ( f_i^{S} , f_i^T )}{\tau_2} \big)}{\text{ exp} \big( \frac{h ( f_i^{S} , f_i^T )}{\tau_2} \big) + \sum_{j \neq i} \text{ exp} \big( \frac{h ( f_i^{S} , f_j^T )}{\tau_2} \big)},
\end{equation}
where $h: \{ d^S, d^T \} \rightarrow \left[ 0, 1 \right]$ is a learnt similarity metric, and $\tau_2$ is a scalar temperature parameter. 
From an information theoretic perspective, the InfoNCE-style loss aligns the teacher and student embedding spaces by maximizing the mutual information among them.
Fig.\ref{fig:contrastive} illustrates G-CRD.

G-CRD generalizes Contrastive Representation Distillation \cite{tian2019contrastive} from sample-level 2D image classification to fine-grained node-level tasks on graphs, and involves key design choices for GNNs:


\begin{figure}[t!]
    \centering
    \includegraphics[width=0.6\linewidth]{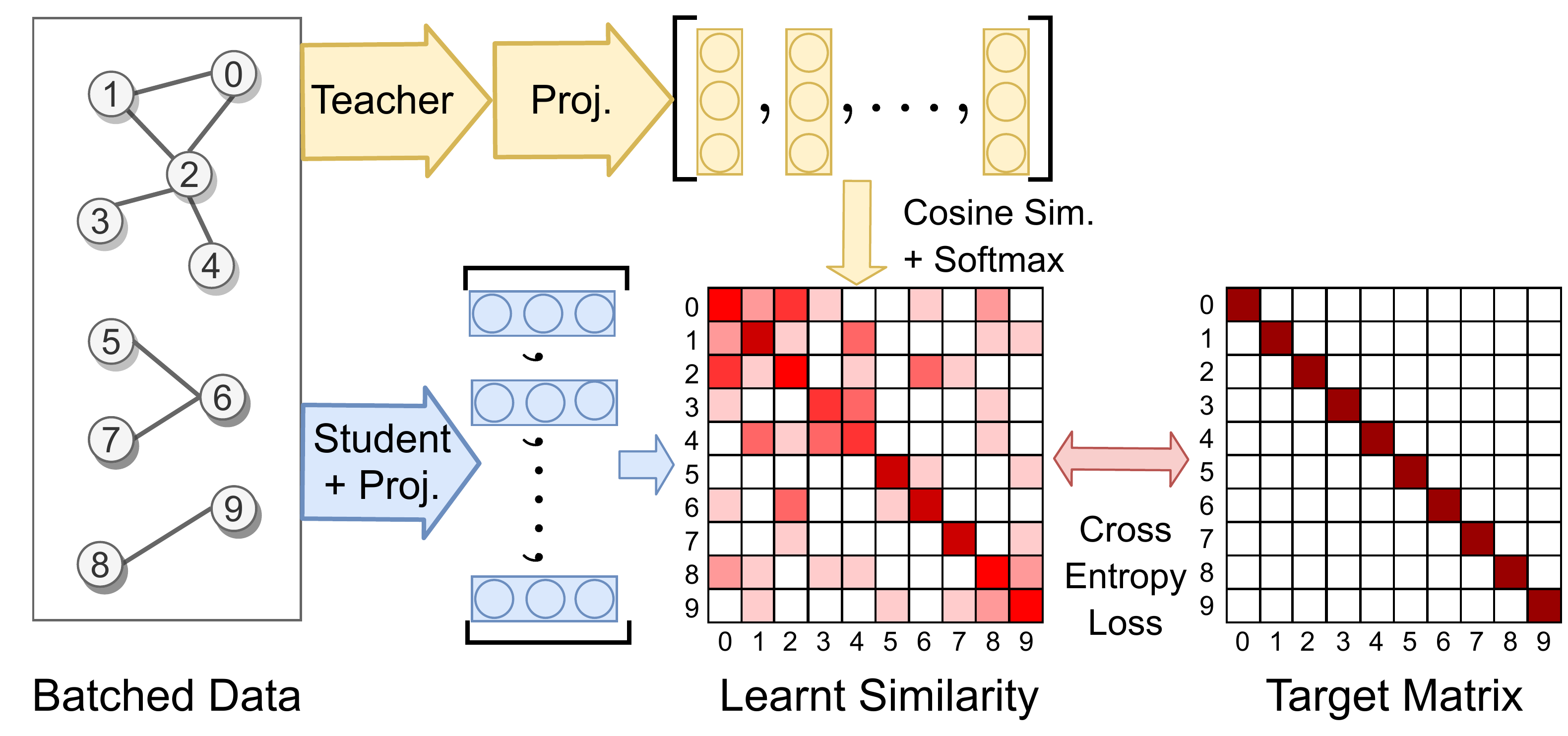}
    \caption{
    \textbf{Graph Contrastive Representation Distillation.} 
    G-CRD is a node-level contrastive learning task of identifying positive node correspondences across the teacher and student embedding spaces, while pushing away a set of distractor nodes. 
    Here, we define negative samples for each student node feature vector as all the other node features vectors from the teacher within the same mini-batch.
    }
    \label{fig:contrastive}
\end{figure}


\textbf{Projection heads.}
We define the similarity metric as a learnt cosine similarity between teacher and student features projected to a common representation space:
\begin{equation}
    \label{eq:G-CRD-sim}
    h( f^S , f^T ) = \frac{P^S \left( f^S \right)' \cdot P^T \left( f^T \right)}{\lVert P^S \left( f^S \right) \rVert \cdot \lVert P^T \left( f^T \right) \rVert},
\end{equation}
where the projection heads $P^S, P^T$ can be multi-layer perceptrons (MLPs) composed of linear transformations to a common dimension $d$ (usually that of the student feature) followed by batch normalization and non-linear activation.
However, independent MLPs on node feature vectors are `structure-agnostic' \cite{sign_icml_grl2020} -- they cannot adapt the shared representation space to the graph structure of the underlying data. 
Thus, we introduce structure-aware projection heads, which use a single GCN layer \cite{kipf2017semi}\footnote{
We chose GCN as it is one of the most well studied and lightweight layers.
} followed by batch normalization and non-linear activation:
\begin{equation}
    \label{eq:G-CRD-proj}
    P \left( f_i \right) =
    \begin{cases}
        \text{GCN} \left( f_i, f_j | (i,j) \in \mathcal{E} \right), & \text{Structure-aware}, \\
        \text{MLP} \left( f_i \right), & \text{Structure-agnostic}, \\
    \end{cases}
\end{equation}
The GCN projection can learn neighborhood-dependent projections via one message passing step, while having the same number of parameters as the structure-agnostic MLP.
We tune the choice of projection head as a hyperparameter. (See App.\ref{app:gcrd} for ablation studies.)

\textbf{Node-level contrastive learning.}
We formulate G-CRD as a node-level contrastive learning task of identifying positive node correspondences across the teacher and student embedding spaces, while pushing away a set of distractors. 
G-CRD adapts the well-studied InfoNCE loss in \eqref{eq:G-CRD}.
Contrastive Representation Distillation (CRD) \cite{tian2019contrastive} is a related representation distillation objective for 2D image classification models:
\begin{eqnarray}
    \label{eq:crd}
    \mathcal{L}_{\text{CRD}} = -\underset{h}{\text{max}} \; \Big( \mathbb{E}_{q(T, S | C=1)} [\log h(T, S)] + \\
    N\mathbb{E}_{q(T, S | C=0)} [1-\log(h(T, S))] \Big), \nonumber
\end{eqnarray}
where $T, S$ are random variables for teacher and student features, $q(T, S | C=1)$ is their joint distribution (positives), $q(T, S | C=0)$ is their marginal distribution (negatives), and $h: \{ d^S, d^T \} \rightarrow \left[ 0, 1 \right]$ is an auxiliary `critic' model which returns an unnormalized similarity score.
Crucially, CRD is tailored for 2D image classification and contrasts among global/per-sample feature vectors.
It additionally requires a specialized memory bank \cite{wu2018unsupervised} to provide a large number $N$ of negative samples.
On the other hand, GNNs build representations at the node-level for both node-level prediction tasks as well as global-level tasks \cite{mesquita2020rethinking}.
Thus, it is not possible to directly apply CRD to GNNs beyond graph-level prediction. 
Our experiments show how a naive application of CRD for GNNs is ineffective, and ablate the impact of our design choices in formulating G-CRD. 

\textbf{Mini-batch negative sampling.}
Contrastive learning among GNN node features from teachers and students alleviates the need for specialized negative sampling: 
we simply define negative samples for each student node feature as all the other node features from the teacher within the same mini-batch.
This is markedly different from recent contrastive pre-training objectives for GNNs, which rely on handcrafted data augmentation \cite{hu2019strategies, you2020graph} or sub-graph sampling procedures \cite{jiao2020subgraph, qiu2020gcc} to generate multiple views of graphs.
Unlike sample-level CRD, our approach does not require extremely large batch sizes or negative samples, as the combined cardinality of all nodes within a mini-batch is sufficient for boosting student performance.
G-CRD is robust across a range of graph and batch sizes -- from 25 nodes per graph in mini-batches of 32, up to single giant graphs with over 1.9 million nodes.

\section{Experimental Setup}

\subsection{Datasets}

We benchmark distillation techniques across a range of tasks on single large-scale networks, batches of small graphs, as well as batches of 3D point clouds, summarized in Tab.\ref{tab:datasets}. 
As we are motivated by real-world and real-time applications involving noisy and shifting data distributions, we make the following considerations for our benchmarks:

\textbf{Challenging for student models.}
Past work on distillation for GNNs \cite{yang2020distilling} used small-scale PPI \cite{hamilton2017inductive} (node classification) and ModelNet \cite{wu20153d} (3D point cloud classification), both of which are considered saturated by the community \cite{hu2020open,uy-scanobjectnn-iccv19}. 
The performance gap between cumbersome teachers and lightweight student models is negligible for these datasets, which makes them unsuitable benchmarks for comparing techniques.
(See App.\ref{app:ppi} for an investigation on PPI.)

We evaluate node classification on ARXIV and the Microsoft Academic Graph (MAG) \cite{wang2020microsoft, hu2020open}, which are 70$\times$ and 800$\times$ larger than PPI, respectively, and consider the more challenging semantic segmentation task on 3D point cloud scenes from S3DIS \cite{armeni20163d} that are over 10$\times$ denser than CAD object scans from ModelNet.
We additionally evaluate graph classification on MOLHIV from OGB/MoleculeNet \cite{wu2018moleculenet}.

\textbf{Out-of-distribution evaluation.}
Unlike PPI and ModelNet, all datasets involve realistic and carefully curated train-test splitting procedures to evaluate out-of-distribution generalization.
ARXIV and MAG follow a temporal split, MOLHIV trains on common molecular scaffolds while testing on rare ones, and S3DIS is split according to 6 distinct areas.


\begin{table*}[t]
  \centering
  \caption{
  \textbf{Summary of datasets.} 
  ARXIV, MAG, and MOLHIV were accessed via OGB \cite{hu2020open} and S3DIS via PyTorch Points 3D \cite{tp3d}. 
  }
  \resizebox{\linewidth}{!}{
  \begin{tabular}{lrrrccccc}
    \toprule
    {\textbf{Name}} & {\textbf{\#Samples}} & \textbf{Avg. \#Nodes} & \textbf{Avg. \#Edges} & \textbf{Split Scheme} & \textbf{Split Ratio} & \textbf{Prediction} & \textbf{Task} & {\textbf{Metric}} \\
    \midrule
    \midrule
    MOLHIV & 41,127 & 25.5 & 27.5 & Scaffold & 80/10/10 & Graph-level & Binary clf. & ROC-AUC \\
    S3DIS & 5,845 & 20,000 & - & Rooms & 52/19/29 & Point-level & Multi-class clf. (16) & mIoU, mAcc \\
    ARXIV & 1 & 169,343 & 1,166,243 & Time & 54/18/28 & Node-level & Multi-class clf. (40) & Accuracy \\
    MAG & 1 & 1,939,743 & 21,111,007 & Time & 85/9/6 & Node-level & Multi-class clf. (349) & Accuracy \\
    \midrule
    PPI & 24 & 2,372 & 34,113 & Random & 84/8/8 & Node-level & Multi-label bin. clf. (128) & F1 \\ 
    \bottomrule
  \end{tabular}
  }
  \label{tab:datasets}
\end{table*}



\subsection{Teacher and Student Architectures}
\label{sec:exp:models}

Our choice of teachers and student architectures are made with the following considerations about deploying GNNs:
(1) \textit{depth} -- more message passing rounds allow models to access larger sub-graphs around each node, at the cost of training and inference time;
(2) \textit{hidden channels} -- especially for giant graphs, increasing hidden channels boosts performance as well as memory consumption, and may require specialized sampling procedures;
and (3) \textit{architectural complexity} -- principled geometric priors \cite{thomas2019KPConv} or attention mechanisms over edges \cite{velickovic2018graph} improve model expressivity but tend to have higher inference latency and memory usage.

Unlike in \cite{yang2020distilling}, we benchmark distillation techniques across \textit{heterogeneous} architecture families with significant mismatch in terms of parameter count, expressive power, and inference time; \textit{e.g.} for social networks, we distil from Graph Attention Networks \cite{velickovic2018graph} to simple GCNs \cite{kipf2017semi}. 
For 3D point clouds, we distil from Kernel Point ConvNet \cite{thomas2019KPConv} teachers to lightweight PointNet++ \cite{qi2017pointnet++} and voxel-based MinkowskiNet \cite{choy20194d}.
We give further rationale for our choice of teacher-student pairs for each experiment in Sec.\ref{sec:results}.
Details on the configurations, inference time and latency of teacher and student models are available in App.\ref{app:models}


\subsection{Training and Evaluation}

Our implementation is built upon PyTorch Geometric \cite{fey2019fast}, PyTorch Points 3D \cite{tp3d}, and Open Graph Benchmark \cite{hu2020open}.
We follow the best practices and guidelines for each dataset.

\textbf{Training.}
Teacher models are trained via the conventional supervised learning paradigm and the final weights for distillation are selected by early stopping.
Student models are trained via the knowledge distillation pipeline described in Sec.\ref{sec:method}.
For the OGB datasets, we follow their example implementations and training setups for both teacher and student models.
For S3DIS, the teacher models are trained for 600 epochs following their original learning rate strategies,
while the student models are trained for 300 epochs and use an exponential learning rate strategy.
We use the conventional data preparation and augmentation procedures for S3DIS via PyTorch Points 3D.

\textbf{Evaluation.}
For the OGB datasets, we use the official evaluators and report the average test performance across 8/10 random seeds
(each teacher-student pair is re-trained for each seed).
For S3DIS, we follow best practices in the literature: all models are trained once, and we report the average mIoU and mAcc over 10 voting runs on the held-out Area-5 test set using the evaluation protocol from PyTorch Points 3D.

\textbf{Baselines and hyperparameters.}
Following Yang et al. \cite{yang2020distilling}, we compare the GNN representation distillation techniques G-CRD, GSP and LSP to logit-based KD \cite{hinton2015distilling} \eqref{eq:kd} as well as FitNet \cite{romero2014fitnets} and Attention Transfer (AT) \cite{zagoruyko2016paying}, two feature mimicking baselines adapted from 2D computer vision which are formulated as: $\mathcal{L}_{\text{Aux}} = \sum_{i \in \mathcal{V}} \lVert P^T(f^T_i) - P^S(f^S_i) \rVert^2_2$ (FitNet uses L2 normalized projection head, AT uses attention mapping).
We tune the loss balancing weights $\alpha, \beta$ in \eqref{eq:kd-sup-aux} for all techniques on the validation set.
For KD, we tune $\alpha \in \{0.8, 0.9\}, \tau_1 \in \{4, 5\}$.
For FitNet, AT, LSP, and GSP, we tune $\beta \in \{100, 1000, 10000\}$ and the kernel in \eqref{eq:lsp-sim} (only LSP, GSP).
For G-CRD, we tune $\beta \in \{0.01, 0.05\}, \tau_2 \in \{0.05, 0.075, 0.1\}$ and the projection head in \eqref{eq:G-CRD-proj}.
When comparing representation distillation methods, we set $\alpha$ to 0 in order to ablate performance, as in \cite{tian2019contrastive}, and reduce $\beta$ by one order of magnitude.


\section{Results}
\label{sec:results}


\begin{table*}[t!]
    \centering
    \caption{
    \textbf{Molecular graph classification on MOLHIV} (metric: ROC-AUC (\%)).
    \textbf{Bold}/\underline{underlined} denote the best/second best performing distillation technique for each column.
    The arrows (\green{$\uparrow$})/(\red{$\downarrow$}) denote performance improvement/regression compared to logit-based Knowledge Distillation (KD).
    We report the average performance and std. across 8 random seeds.
    }
    \resizebox{\linewidth}{!}{
    \begin{tabular}{llccccc}
        \toprule
        \multicolumn{2}{c}{\textbf{Teacher} (\#Layer,\#Param):} & \textbf{GIN-E} (5L,3.3M) & \textbf{PNA} (5L,2.4M) & \textbf{GIN-E} (5L,3.3M) & \textbf{PNA} (5L,2.4M) & \textbf{PNA} (5L,2.4M) \\
        \multicolumn{2}{c}{\textbf{Student} (\#Layer,\#Param):} & \textbf{GCN} (2L,15K) & \textbf{GCN} (2L,15K) & \textbf{GCN} (2L,40K) & \textbf{GCN} (2L,40K) & \textbf{GIN} (2L,10K) \\
        \midrule
        \midrule
        \multirow{2}{*}{\rotatebox[origin=c]{90}{\text{Sup.}}} 
        & Supervised Teacher & 77.69 \text{$\pm$1.61} & 77.48 \text{$\pm$1.71} & 77.69 \text{$\pm$1.61} & 77.48 \text{$\pm$1.71} & 77.48 \text{$\pm$1.71} \\
        & Supervised Student & 73.02 \text{$\pm$1.46} & 73.02 \text{$\pm$1.46} & 73.65 \text{$\pm$1.50} & 73.65 \text{$\pm$1.50} & 73.03 \text{$\pm$2.02} \\
        \midrule
        \multirow{6}{*}{\rotatebox[origin=c]{90}{\text{Distillation}}}
        & KD \cite{hinton2015distilling} & \underline{74.08} \text{$\pm$1.03} & \underline{74.13} \text{$\pm$1.72} & \underline{75.25} \text{$\pm$1.71} & 74.45 \text{$\pm$1.27} & 73.42 \text{$\pm$2.14} \\
        & FitNet \cite{romero2014fitnets} & 73.62 \text{$\pm$1.05} (\red{$\downarrow$}) & 73.65 \text{$\pm$1.25} (\red{$\downarrow$}) & 74.52 \text{$\pm$1.33} (\red{$\downarrow$}) & 74.39 \text{$\pm$1.46} (\red{$\downarrow$}) & 72.88 \text{$\pm$0.89} (\red{$\downarrow$}) \\
        & AT \cite{zagoruyko2016paying} & 73.85 \text{$\pm$0.85} (\red{$\downarrow$}) & 73.64 \text{$\pm$1.50} (\red{$\downarrow$}) & 74.94 \text{$\pm$0.97} (\red{$\downarrow$}) & 73.89 \text{$\pm$1.92} (\red{$\downarrow$}) & \underline{73.87} \text{$\pm$2.28} (\green{$\uparrow$}) \\
        & LSP \cite{yang2020distilling} & 73.58 \text{$\pm$1.29} (\red{$\downarrow$}) & 73.24 \text{$\pm$1.67} (\red{$\downarrow$}) & 75.04 \text{$\pm$1.20} (\red{$\downarrow$}) & 74.43 \text{$\pm$1.58} (\red{$\downarrow$}) & 70.74 \text{$\pm$1.82} (\red{$\downarrow$}) \\
        & GSP & 72.83 \text{$\pm$1.30} (\red{$\downarrow$}) & 73.74 \text{$\pm$0.93} (\red{$\downarrow$}) & 75.12 \text{$\pm$1.27} (\red{$\downarrow$}) & \underline{75.09} \text{$\pm$1.48} (\green{$\uparrow$}) & 69.68 \text{$\pm$2.88} (\red{$\downarrow$}) \\
        & G-CRD (Ours) & \textbf{74.34} \text{$\pm$1.44} (\green{$\uparrow$}) & \textbf{75.11} \text{$\pm$0.73} (\green{$\uparrow$}) & \textbf{75.53} \text{$\pm$1.64} (\green{$\uparrow$}) & \textbf{75.89} \text{$\pm$0.80} (\green{$\uparrow$}) & \textbf{75.77} \text{$\pm$2.02} (\green{$\uparrow$}) \\
        
        \bottomrule
    \end{tabular}
    }
    \label{tab:molecules}
\end{table*} 



\begin{table*}[th!]
    \centering
    \caption{
    \textbf{3D Semantic segmentation on S3DIS} (metric: mIoU, mAcc).
    \textbf{Bold}/\underline{underlined} denote the best/second best performing distillation technique for each column.
    The arrows (\green{$\uparrow$})/(\red{$\downarrow$}) denote performance improvement/regression compared to logit-based Knowledge Distillation (KD).
    We report the average performance over 10 voting runs.
    }
    \resizebox{\linewidth}{!}{
    \begin{tabular}{llcccc}
        \toprule
        \multicolumn{2}{c}{\textbf{Teacher} (\#Param):} & \textbf{KP-FCNN} (14.0M) & \textbf{SPVCNN} (21.8M) & \textbf{KP-FCNN} (14.0M) & \textbf{SPVCNN} (21.8M) \\
        \multicolumn{2}{c}{\textbf{Student} (\#Param):} & \textbf{PointNet, SSG} (1.4M) & \textbf{PointNet, SSG} (1.4M) & \textbf{MinkNet, 20\%cr.} (0.8M) & \textbf{MinkNet, 20\%cr.} (0.8M) \\
        \midrule
        \midrule
        \multirow{2}{*}{\rotatebox[origin=c]{90}{\text{Sup.}}}
        & Supervised Teacher & 62.70, 69.54 & 64.58, 71.71 & 62.70, 69.54 & 64.58, 71.71 \\
        & Supervised Student & 49.67, 57.51 & 49.67, 57.51 &  55.29, 64.14 & 55.29, 64.14 \\
        \midrule
        \multirow{6}{*}{\rotatebox[origin=c]{90}{\text{Distillation}}}
        & KD \cite{hinton2015distilling} & 51.89, 59.48 & \underline{51.81, 59.22} & 56.00, 64.90 & \underline{55.78, 64.51} \\
        & FitNet \cite{romero2014fitnets} & 49.37, 57.35 (\red{$\downarrow$}) & 49.85, 57.78 (\red{$\downarrow$}) & 48.94, 57.26  (\red{$\downarrow$}) & 53.62, 63.14 (\red{$\downarrow$}) \\
        & AT \cite{zagoruyko2016paying} & 51.82, 59.38 (\red{$\downarrow$}) & 49.57, 57.13 (\red{$\downarrow$}) & \underline{56.02, 64.87} (\green{$\uparrow$}) & 54.84, 63.78 (\red{$\downarrow$}) \\
        & LSP \cite{yang2020distilling} & 50.69, 58.49 (\red{$\downarrow$}) & 51.07, 58.57 (\red{$\downarrow$}) & 54.20, 63.59  (\red{$\downarrow$}) & 55.20, 64.34 (\red{$\downarrow$}) \\
        & GSP & \underline{53.00, 61.15} (\green{$\uparrow$}) & 51.68, 60.35 (\red{$\downarrow$}) & 55.50, 65.19  (\red{$\downarrow$}) & 54.77, 63.88 (\red{$\downarrow$}) \\
        & G-CRD (Ours) & \textbf{53.15, 61.15} (\green{$\uparrow$}) & \textbf{53.27, 61.29} (\green{$\uparrow$}) & \textbf{56.07, 64.87} (\green{$\uparrow$}) & \textbf{55.83, 65.03} (\green{$\uparrow$}) \\
        \bottomrule
    \end{tabular}
    }
    \label{tab:s3dis}
\end{table*} 


\subsection{Molecular Graph Property Prediction}

We consider the graph-level property prediction task over batches of molecular graphs.
Reducing the inference latency of GNNs for molecules speeds up high throughput virtual screening \cite{wu2018moleculenet}.
Additionally, virtual screening on proprietary data will require homomorphically encrypted models, which further demand low layer count and hidden size \cite{qaisarahmadalbadawi2020towards}.
As teachers, we consider 5-layer deep GIN-E \cite{hu2019strategies} and PNA \cite{corso2020principal} augmented with virtual nodes, two strong OGB baselines.
Notably, PNA is the most expressive message passing GNN but explicitly materializes messages over graph edges, leading to higher memory requirement and inference latency.
Our student architectures are 2-layer GCN \cite{kipf2017semi} and GIN \cite{xu2018how}, which are comparatively less expressive and do not use virtual nodes (and edge features in GIN), while having low inference latency.

In Tab.\ref{tab:molecules}, we compare logit-based KD and the representation distillation techniques across a range of teacher-student pairs.
\textbf{We find that the implicit G-CRD technique consistently improves over the supervised student's performance and outperforms the explicit global structure preserving approach, GSP. In turn, GSP outperforms the purely local approach, LSP.}
While all other distillation techniques do offer minor performance boosts over the supervised student, G-CRD is the only one which improves over the KD baseline for most teacher-student combinations.


\subsection{3D Point Cloud Semantic Segmentation}

Semantic segmentation of 3D scene graphs is a safety critical real-time task with applications in autonomous driving and robotics.
Models process 3D scenes as sets of points or voxel grids, and make a dense prediction by assigning semantic categories to each point/unit.
Recent state-of-the-art architectures involve strong geometric priors and complex architectures, leading to increased inference latency and GPU memory requirement.
Here, we consider distilling two powerful models, Kernel Point Convolution (KP-FCNN, rigid kernel) \cite{thomas2019KPConv} and Sparse Point-Voxel CNN (SPVCNN) \cite{liu2019point, tang2020searching}, into simpler models with low inference latency and memory usage: PointNet++ with single-scale grouping \cite{qi2017pointnet++} and voxel-based MinkowskiNet \cite{choy20194d} at low 20\% channel ratio.

In Tab.\ref{tab:s3dis}, we see trends that are consistent with the previous section: G-CRD consistently improves the performance of students compared to GSP, and is particularly effective for boosting PointNet++.
\textbf{Notably, due to a large mismatch in teacher-student representation capacity, feature mimicking losses, FitNet and AT, may worsen the student's performance over purely supervised learning.}


\subsection{Node Classification on Social Networks}


\begin{table*}[t!]
    \centering
    \caption{
    \textbf{Node classification on ARXIV and MAG} (metric: Accuracy (\%)).
    \textbf{Bold}/\underline{underlined} denote the best/second best performing distillation technique for each column.
    The arrows (\green{$\uparrow$})/(\red{$\downarrow$}) denote performance improvement/regression compared to logit-based Knowledge Distillation (KD).
    We report the average performance and std. across 10 random seeds.
    }
    \resizebox{\linewidth}{!}{
    \begin{tabular}{llcccc}
        \toprule
        \multicolumn{2}{r}{\textbf{Dataset}:} & ARXIV & ARXIV & ARXIV & MAG \\
        \multicolumn{2}{c}{\textbf{Teacher} (\#Layer,\#Param):} & \textbf{GAT} (3L,1.4M) & \textbf{GAT} (3L,1.4M) & \textbf{GAT} (3L,1.4M) & \textbf{R-GCN} (3L,5.5M) \\
        \multicolumn{2}{c}{\textbf{Student} (\#Layer,\#Param):} & \textbf{GCN} (2L,44K) & \textbf{GraphSage} (2L,87K) & \textbf{SIGN} (3L,3.5M) & \textbf{R-GCN} (2L,170K) \\
        \midrule
        \midrule
        \multirow{2}{*}{\rotatebox[origin=c]{90}{\text{Sup.}}}
        & Supervised Teacher & 73.91 \text{$\pm$0.12} & 73.91 \text{$\pm$0.12} & 73.91 \text{$\pm$0.12} & 49.48 \text{$\pm$0.35} \\
        & Supervised Student & 71.25 \text{$\pm$0.28} & 70.97 \text{$\pm$0.23} & 71.98 \text{$\pm$0.16} & 46.22 \text{$\pm$0.31} \\
        \midrule
        \multirow{6}{*}{\rotatebox[origin=c]{90}{\text{Distillation}}}
        & KD \cite{hinton2015distilling} & \underline{71.55} \text{$\pm$0.25} & \textbf{71.44} \text{$\pm$0.10} & \textbf{72.26} \text{$\pm$0.11} & \textbf{46.65} \text{$\pm$0.20} \\
        & FitNet \cite{romero2014fitnets} &  71.38 \text{$\pm$0.17} (\red{$\downarrow$}) & 70.78 \text{$\pm$0.25} (\red{$\downarrow$}) & 71.98 \text{$\pm$0.13} (\red{$\downarrow$}) & 46.15 \text{$\pm$0.24} (\red{$\downarrow$}) \\
        & AT \cite{zagoruyko2016paying} & 70.44 \text{$\pm$0.28} (\red{$\downarrow$}) & 70.17 \text{$\pm$0.11} (\red{$\downarrow$}) & 71.99 \text{$\pm$0.14} (\red{$\downarrow$}) & 46.09 \text{$\pm$0.27} (\red{$\downarrow$}) \\
        & LSP \cite{yang2020distilling} & 71.52 \text{$\pm$0.22} (\red{$\downarrow$}) & 70.95 \text{$\pm$0.22} (\red{$\downarrow$}) & - & 46.23 \text{$\pm$0.41} (\red{$\downarrow$}) \\
        & GSP & 71.41 \text{$\pm$0.31} (\red{$\downarrow$}) & 70.98 \text{$\pm$0.33} (\red{$\downarrow$}) &  71.99 \text{$\pm$0.12} (\red{$\downarrow$}) & 46.04 \text{$\pm$0.15} (\red{$\downarrow$}) \\
        & G-CRD (Ours) & \textbf{71.64} \text{$\pm$0.16} (\green{$\uparrow$}) & \underline{71.15} \text{$\pm$0.12} (\red{$\downarrow$}) & \underline{72.10} \text{$\pm$0.10} (\red{$\downarrow$}) & \underline{46.42} \text{$\pm$0.20} (\red{$\downarrow$}) \\
        \midrule
        \multirow{5}{*}{\rotatebox[origin=c]{90}{\text{KD + Dist.}}}
        & KD + FitNet \cite{romero2014fitnets} & 71.10 \text{$\pm$0.19} (\red{$\downarrow$}) & 71.11 \text{$\pm$0.17} (\red{$\downarrow$}) & \underline{72.31} \text{$\pm$0.08} (\green{$\uparrow$}) & 46.60 \text{$\pm$0.36} (\red{$\downarrow$}) \\
        & KD + AT \cite{zagoruyko2016paying} & 70.91 \text{$\pm$0.31} (\red{$\downarrow$}) & 71.06 \text{$\pm$0.19} (\red{$\downarrow$}) & 72.27 \text{$\pm$0.11} (\green{$\uparrow$}) & 46.59 \text{$\pm$0.41} (\red{$\downarrow$}) \\
        & KD + LSP \cite{yang2020distilling} & 71.35 \text{$\pm$0.23} (\red{$\downarrow$}) & 71.34 \text{$\pm$0.17} (\red{$\downarrow$}) & - & \underline{46.73} \text{$\pm$0.31} (\green{$\uparrow$}) \\
        & KD + GSP & 71.39 \text{$\pm$0.19} (\red{$\downarrow$}) & \underline{71.51} \text{$\pm$0.18} (\green{$\uparrow$}) & 72.27 \text{$\pm$0.16} (\green{$\uparrow$}) & 46.49 \text{$\pm$0.18} (\red{$\downarrow$}) \\
        & KD + G-CRD (Ours) & \textbf{71.57} \text{$\pm$0.23} (\green{$\uparrow$}) & \textbf{71.59} \text{$\pm$0.15} (\green{$\uparrow$}) & \textbf{72.32} \text{$\pm$0.11} (\green{$\uparrow$}) & \textbf{46.78} \text{$\pm$0.32} (\green{$\uparrow$}) \\
        \bottomrule
    \end{tabular}
    }
    \label{tab:arxiv}
\end{table*} 


Reducing model depth and feature dimensions boosts inference and reduces memory usage for real-time applications on large-scale graphs. 
For ARXIV, a homogeneous citation network where models classify the subject of each node/paper, we use a strong 3-layer GAT \cite{velickovic2018graph} teacher.
GAT has high memory requirement during training and inference due to its attention mechanism.
We distil into 2-layer GCN \cite{kipf2017semi} and GraphSage \cite{hamilton2017inductive}, which are comparatively more scalable but lack the expressivity of attention.
We also distil to structure-agnostic SIGN \cite{sign_icml_grl2020} designed for parallelized large-graph processing.
LSP cannot be used for SIGN as it does not make use of graph structure.

Experiments on the giant heterogeneous Microsoft Academic Graph (MAG) of authors, papers, and institutions further tests the scalability of distillation.
We distil among Relational-GCNs \cite{schlichtkrull2018modeling} at different depth and feature sizes.
We use the GraphSAINT \cite{Zeng2020GraphSAINT:} sampler to fit GPU memory.

In Tab.\ref{tab:arxiv}, when considering each technique independently, logit-based KD is the best while G-CRD is the second best.
We believe this can be explained by the homophily phenomenon for node labels on social networks \cite{huang2021combining}.
The soft labels from the teacher model thus providing a more informative signal for knowledge transfer than the latent representations.
On combining KD with representation distillation, KD + G-CRD consistently outperforms all other pairs.
\textbf{GSP is the worst performing technique for MAG, demonstrating the inability of explicit global topology preservation to scale to large graphs.}
Note that performance gains from distillation may seem marginal, but are significant as metrics are averaged over several thousand nodes (48K for ARXIV, 42K for MAG).




\subsection{Does Distilation Preserve Topology?}

Across Tab.\ref{tab:molecules}, \ref{tab:arxiv}, \ref{tab:s3dis}, we have observed that G-CRD, which implicitly preserves global topology from the teacher to the student embedding space, outperforms explicit structure preserving approaches LSP and GSP, as well as feature mimicking baselines FitNet and AT.
In order to unpack the efficacy of G-CRD beyond performance metrics, we measure how well distillation techniques preserve both global and local topology from the teacher to the student node embedding space in Tab.\ref{tab:arxiv-correlation}.
To quantify global representational similarity between the teacher and student, we use the recently proposed Centered Kernel Alignment score (CKA) \cite{kornblith2019similarity,nguyen2021wide} between two embedding sets, as well as the classical Mantel Test \cite{mantel1967detection,legendre2012numerical}\footnote{
We use the implementation from scikit-bio (skbio.stats.distance.mantel).
} of Pearson correlation between all pairwise cosine distances from two embedding sets (as in the GSP loss).
Additionally, we quantify local structural similarity via another Mantel Test which only considers distances over pre-defined edges (as in the LSP loss).

Our results largely follow the intuitions developed in Fig.\ref{fig:main}:
In terms of global topology and representational similarity, students trained with FitNet, GSP, and G-CRD have high correlations to the teacher.
On the other hand, LSP is the most correlated for local topology over existing graph edges, but relatively poorer at preserving global topology.
\textbf{Overall, we see that the implicit contrastive approach G-CRD strikes a balance between preserving both local and global relationships, while the explicit LSP/GSP are best at preserving only one or the other.}


\begin{table*}[t!]
    \centering
    \caption{
    \textbf{Topological similarity among teacher and student embeddings on ARXIV}.
    We report average metrics across 10 random seeds for teacher and student validation set node embeddings.
    CKA Score (range: $[0, 1]$) measures global representational similarity between embedding sets.
    Global Mantel Test (range: $[-1, 1]$) measures Pearson correlation between the pairwise cosine distance matrices of embedding sets, \textit{i.e.} all pairwise relationships or global structural similarity.
    Local Mantel Test measures local structural similarity by only considering distances over pre-defined edges.  
    Colors denote the \textbf{highest}/\red{\textbf{second}}/\blue{\textbf{third}} highest similarity.
    }
    \resizebox{\linewidth}{!}{
    \begin{tabular}{lcccccc}
        \toprule
        \textbf{Similarity Metric:} & \multicolumn{2}{c}{\textbf{CKA Score} \cite{kornblith2019similarity}} & \multicolumn{2}{c}{\textbf{Global Mantel Test} \cite{mantel1967detection}} & \multicolumn{2}{c}{\textbf{Local Mantel Test} \cite{mantel1967detection}} \\
        \textbf{Teacher} (\#Layer,\#Param): & \textbf{GAT} (3L,1.4M) & \textbf{GAT} (3L,1.4M) & \textbf{GAT} (3L,1.4M) & \textbf{GAT} (3L,1.4M) & \textbf{GAT} (3L,1.4M) & \textbf{GAT} (3L,1.4M) \\
        \textbf{Student} (\#Layer,\#Param): & \textbf{GCN} (2L,44K) & \textbf{GraphSage} (2L,87K) & \textbf{GCN} (2L,44K) & \textbf{GraphSage} (2L,87K) & \textbf{GCN} (2L,44K) & \textbf{GraphSage} (2L,87K) \\
        \midrule
        \midrule
        Sup. Teacher & 1.000 \text{$\pm$0.000} & 1.000 \text{$\pm$0.000} & 1.000 \text{$\pm$0.000} & 1.000 \text{$\pm$0.000} & 1.000 \text{$\pm$0.000} & 1.000 \text{$\pm$0.000} \\
        Sup. Student & 0.655 \text{$\pm$0.011} & 0.609 \text{$\pm$0.007} & 0.680 \text{$\pm$0.011} & 0.623 \text{$\pm$0.008} & 0.695 \text{$\pm$0.005} & 0.521 \text{$\pm$0.007} \\
        \midrule
        KD \cite{hinton2015distilling} & 0.716 \text{$\pm$0.011} & 0.721 \text{$\pm$0.003} & 0.733 \text{$\pm$0.007} & 0.730 \text{$\pm$0.004} & 0.710 \text{$\pm$0.006} & 0.589 \text{$\pm$0.007} \\
        KD + FitNet \cite{romero2014fitnets} & \red{\textbf{0.740}} \text{$\pm$0.006} & \textbf{0.764} \text{$\pm$0.005} & \textbf{0.760} \text{$\pm$0.005} & \textbf{0.765} \text{$\pm$0.005} & \blue{\textbf{0.736}} \text{$\pm$0.008} & 0.576 \text{$\pm$0.009} \\
        KD + AT \cite{zagoruyko2016paying} & 0.570 \text{$\pm$0.020} & 0.696 \text{$\pm$0.013} & 0.593 \text{$\pm$0.015} & 0.687 \text{$\pm$0.012} & 0.669 \text{$\pm$0.009} & 0.478 \text{$\pm$0.007} \\
        KD + LSP \cite{yang2020distilling} & 0.714 \text{$\pm$0.006} & 0.708 \text{$\pm$0.005} & 0.697 \text{$\pm$0.011} & 0.683 \text{$\pm$0.005} & \textbf{0.772} \text{$\pm$0.006} &  \textbf{0.648} \text{$\pm$0.008} \\
        KD + GSP & \textbf{0.746} \text{$\pm$0.010} & \red{\textbf{0.752}} \text{$\pm$0.004} & \red{\textbf{0.756}} \text{$\pm$0.006} & \red{\textbf{0.757}} \text{$\pm$0.003} & 0.722 \text{$\pm$0.007} & \blue{\textbf{0.594}} \text{$\pm$0.009} \\
        KD + G-CRD (Ours) & \blue{\textbf{0.725}} \text{$\pm$0.007} & \blue{\textbf{0.727}} \text{$\pm$0.003} & \blue{\textbf{0.750}} \text{$\pm$0.004} & \blue{\textbf{0.753}} \text{$\pm$0.003} & \red{\textbf{0.742}} \text{$\pm$0.004} & \red{\textbf{0.596}} \text{$\pm$0.006} \\
        \bottomrule
    \end{tabular}
    }
    \label{tab:arxiv-correlation}
\end{table*} 




\begin{table}[t!]
    \centering
    \caption{
    \textbf{Molecular graph classification on MOLHIV} under INT8 quantization (metric: ROC-AUC (\%)).
    }
    \resizebox{0.6\linewidth}{!}{
    \begin{tabular}{llcc}
        \toprule
        \multicolumn{2}{c}{\textbf{Teacher} (\#Param):} & \textbf{PNA} @FP32 (2.4M) & \textbf{PNA} @FP32 (2.4M) \\
        \multicolumn{2}{c}{\textbf{Student} (\#Param):} & \textbf{GIN} @INT8 (10K) & \textbf{GIN} @INT8 (10K) \\
        \multicolumn{2}{c}{\textbf{Training Scheme:}} & Vanilla QAT & DegreeQuant \cite{tailor2021degreequant} \\
        \midrule
        \midrule
        \multirow{2}{*}{\rotatebox[origin=c]{90}{\text{Sup.}}}
        & Teacher & 77.48 \text{$\pm$1.71} & 77.48 \text{$\pm$1.71} \\
        & Student & 70.62 \text{$\pm$3.05} & 71.38 \text{$\pm$2.38} \\
        \midrule
        \multirow{6}{*}{\rotatebox[origin=c]{90}{\text{Distillation}}}
        & KD \cite{hinton2015distilling} & 72.93 \text{$\pm$1.14} & \underline{71.74} \text{$\pm$1.97} \\
        & FitNet \cite{romero2014fitnets} & \underline{73.03} \text{$\pm$1.89} (\green{$\uparrow$}) & 71.08 \text{$\pm$3.86} (\red{$\downarrow$}) \\
        & AT \cite{zagoruyko2016paying} & 72.76 \text{$\pm$1.02} (\red{$\downarrow$}) & 69.31 \text{$\pm$2.12} (\red{$\downarrow$}) \\
        & LSP \cite{yang2020distilling} & 70.91 \text{$\pm$2.23} (\red{$\downarrow$}) & 68.94 \text{$\pm$4.23} (\red{$\downarrow$}) \\
        & GSP & 66.44 \text{$\pm$4.06} (\red{$\downarrow$}) & 68.84 \text{$\pm$2.29} (\red{$\downarrow$}) \\
        & G-CRD (Ours) & \textbf{73.65} \text{$\pm$2.09} (\green{$\uparrow$}) & \textbf{73.50} \text{$\pm$1.13} (\green{$\uparrow$}) \\
        \bottomrule
    \end{tabular}
    }
    \label{tab:molecules-quant}
\end{table}


\subsection{Quantizability of Distilled Representations}

Quantization of model weights and activations to lower precision arithmetic is a complementary compression technique to distillation.
Models at lower precision such as 8-bit integers have significantly faster inference latency and lower memory usage.
However, quantization aware training (QAT) is known to degrade performance. 
In Tab.\ref{tab:molecules-quant}, we explore whether distillation can improve the performance of a lightweight quantized 2-layer GIN model when trained with vanilla QAT as well as DegreeQuant \cite{tailor2021degreequant}, a GNN-specific QAT.
We find that QAT with G-CRD enables quantized models to retain a large portion of their performance at 8-bit integer precision as compared to other distillation techniques.
Significantly, GIN trained with G-CRD at INT8 (73.65\% ROC-AUC) can still outperform its purely supervised counterpart at full 32-bit floating point precision (73.03\%, see column 6 in Tab.\ref{tab:molecules}).

G-CRD is well suited for QAT because the projection heads ensure that contrastive distillation take place at full precision even when the student model is at low INT8 (the heads do not interfere with the inference phase, which always uses INT8).
On the other hand, structure preserving approaches LSP and GSP are particularly ill suited as they match relationships from low INT8 feature vectors (student) to full precision (teacher). 


\subsection{Transferability of Distilled Representations}

In biomedical discovery, we are interested in models' ability to generalize or extrapolate to unseen regions of chemical space.
In addition to evaluating on out-of-distribution molecular scaffolds from MOLHIV's test set, we further test the transferability of distilled models on smaller scale molecule datasets from OGB: MOLBACE (1,513 samples), MOLSIDER (1,427 samples), and MOLESOL (1,128 samples) in Tab.\ref{tab:transferability}.
We perform linear probing on frozen node features from a 2-layer GCN model.
The GCN feature extractor can either be initialized randomly, pre-trained on MOLHIV via supervised learning, or pre-trained via various distillation techniques with a 5-layer PNA teacher.
Encouragingly, we find that distillation leads to more transferable representations than random or purely supervised initialization.
Overall, G-CRD boosts the transferability of student models for 2 out of 3 datasets with structurally different molecules as well as task semantics.


\begin{table}[t!]
    \centering
    \caption{
    \textbf{Molecular graph classification/regression on small-scale MOL* datasets} (metric: ROC-AUC (\%) for BACE, SIDER; MSE for ESOL).
    We perform linear probing on frozen node representations from GCN (2L,40K) initialized randomly or pre-trained on MOLHIV via supervised learning or distillation techniques.
    }
    \resizebox{0.65\linewidth}{!}{
    \begin{tabular}{lccc}
         \toprule
         Initialization & MOLBACE & MOLSIDER & MOLESOL \\
         \midrule \midrule
         Random & 67.03 \text{$\pm$1.63} & 55.35 \text{$\pm$1.05} & 2.097 \text{$\pm$0.128} \\
         Supervised & 71.32 \text{$\pm$2.48} & 58.40 \text{$\pm$1.18} & 1.907 \text{$\pm$0.064} \\
         \midrule
         KD \cite{hinton2015distilling} & \underline{72.24} \text{$\pm$2.38} & \textbf{59.50} \text{$\pm$0.73} & 1.948 \text{$\pm$0.084} \\
         FitNet \cite{romero2014fitnets} & 71.82 \text{$\pm$4.17} (\red{$\downarrow$}) & 57.31 \text{$\pm$1.13} (\red{$\downarrow$}) & 2.032 \text{$\pm$0.137} (\red{$\downarrow$}) \\
         AT \cite{zagoruyko2016paying} & 69.73 \text{$\pm$3.57} (\red{$\downarrow$}) & 57.62 \text{$\pm$0.43} (\red{$\downarrow$}) & \underline{1.820} \text{$\pm$0.062} (\green{$\uparrow$}) \\
         LSP \cite{yang2020distilling} & 70.03 \text{$\pm$2.12} (\red{$\downarrow$}) & \underline{58.67} \text{$\pm$0.71} (\red{$\downarrow$}) & 1.998 \text{$\pm$0.058} (\red{$\downarrow$}) \\
         GSP & 70.53 \text{$\pm$1.74} (\red{$\downarrow$}) & 58.57 \text{$\pm$1.13} (\red{$\downarrow$}) & 1.948 \text{$\pm$0.070} (\green{$\uparrow$}) \\
         G-CRD (Ours) & \textbf{72.46} \text{$\pm$2.19} (\green{$\uparrow$}) & 56.81 \text{$\pm$1.42} (\red{$\downarrow$}) & \textbf{1.812} \text{$\pm$0.112} (\green{$\uparrow$}) \\
         \bottomrule
    \end{tabular}
    }
    \label{tab:transferability}
\end{table}


\begin{figure}[t!]
 \subfloat[Sparse scans]{
	\begin{minipage}[c][]{0.28\linewidth}
	   \centering
	   \includegraphics[width=1\textwidth]{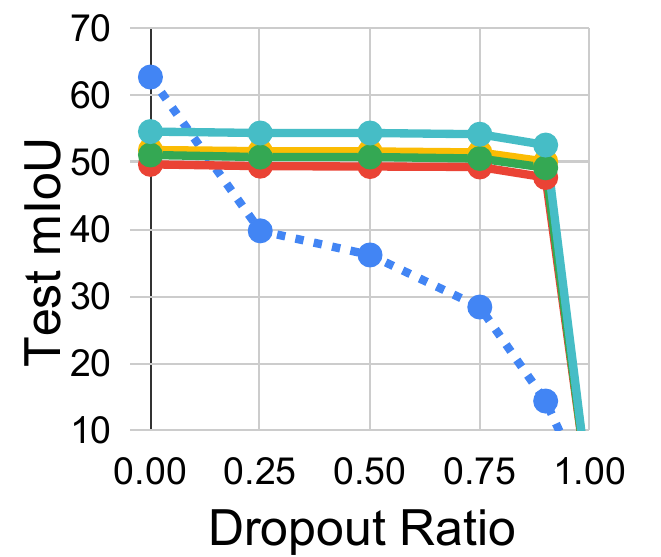}
	\end{minipage}
	}
 \hfill
 \subfloat[Occluded scans]{
	\begin{minipage}[c][]{0.28\linewidth}
	   \centering
	   \includegraphics[width=1\textwidth]{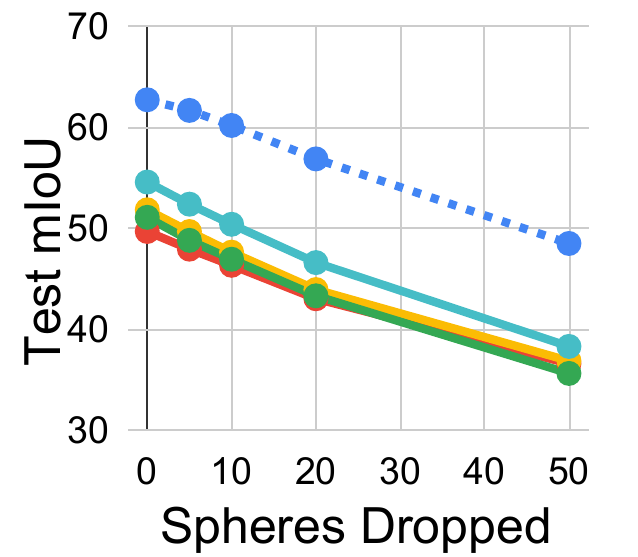}
	\end{minipage}
	}
 \hfill
 \subfloat[Noisy scans]{
	\begin{minipage}[c][]{0.4\linewidth}
	   \centering
	   \includegraphics[width=\textwidth]{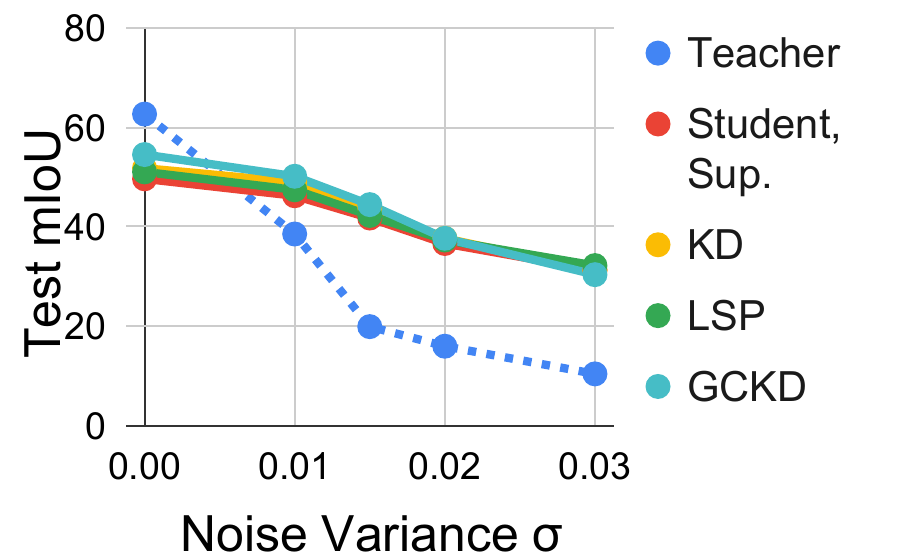}
	\end{minipage}
	}
\caption{
\textbf{Robustness analysis for 3D semantic segmentation.}
}
\label{fig:robustness}
\end{figure}



\subsection{Robustness of Distilled Representations}

Beyond clean test set performance, lightweight 3D segmentation models often have to deal with `dirty' data such as sparse, occluded or noisy scans.
In Fig.\ref{fig:robustness}, we evaluate the impact of distillation on the robustness of PointNet++ across three common challenging scenarios (we also show the KP-FCNN teacher for comparison):
(1) \textit{Sparse scans} -- randomly dropout a percentage of points for each scan;
(2) \textit{Partial and occluded scans} -- dropout all points sampled within a number of random spheres of fixed radius of 0.5m for each scan;
and (3) \textit{Noisy scans} -- add independent random noise to the 3D coordinates of each point with a variance factor $\sigma$.

Training the lightweight student with G-CRD consistently improves its robustness compared to purely supervised training as well as logit-based KD \cite{hinton2015distilling} and LSP \cite{yang2020distilling}.
Promisingly, lightweight students can be more robust to sparse or noisy scans than cumbersome teachers.


\section{Conclusion}
\label{sec:conclusion}

In this work, we study representation distillation for GNNs by training lightweight models to preserve global topology of embeddings from more expressive teacher models. 
We introduce Graph Contrastive Representation Distillation (G-CRD), the first contrastive distillation technique specialized for GNNs.
G-CRD uses contrastive learning to implicitly align the student node embeddings to those of the teacher, preserving structural relationships among graph edges as well as latent interactions among disconnected nodes.

Additionally, we introduce an expanded set of benchmarks on large-scale real-world datasets where the performance gap between teacher and student GNNs is non-negligible.
This was missing from the LSP study which used synthetic and saturated datasets.
Our experiments reveal that training lightweight GNN models with G-CRD consistently improves their performance, robustness, and quantizability compared to explicit structure preserving approaches (LSP, GSP) as well as baselines adapted from 2D computer vision.
We further unpack the efficacy of G-CRD over GSP and LSP through the lens of representational similarity of teacher and student embedding spaces.

Currently, distillation techniques for GNNs are neither as effective nor as well understood as their counterparts in 2D computer vision. 
We hope that our techniques and benchmarks can form the basis for comparison in future work on this emerging research direction.


\section*{Acknowledgements}
This research is supported by the Agency for Science, Technology and Research (A*STAR) under its AME Programmatic Funds (Project No.A19E3b0099 and Project No. A20H6b0151).
We would like to thank Efe Camci, Vijay Prakash Dwivedi, Yoon Ji Wei, Edwin Khoo, Hannes Stärk, Shyam A. Tailor and Wanyue Zhang for helpful comments and discussions.


{
\small
\bibliographystyle{IEEEtran}
\bibliography{IEEEabrv,references}
}


\appendix

\section{Related Work}
\label{sec:related}

\textbf{Graph Neural Networks.}
GNNs \cite{kipf2017semi, hamilton2017inductive, wang2019dynamic, velickovic2018graph, battaglia2018relational} generalize convolutional networks from 2D computer vision to graph structured data.
GNNs serve as powerful feature extractors for real-world and real-time applications across diverse domains, including social networks \cite{ying2018graph, monti2019fake}, biomedicine \cite{stokes2020deep, gainza2020deciphering, long2020predicting}, 3D perception \cite{wang2019dynamic, li2019deepgcns}, natural language processing \cite{joshi2020transformers, ghosal2019dialoguegcn}, and operations research \cite{joshi2019efficient,cappart2021combinatorial}. 
Our study complements an emerging line of work on efficient GNNs \cite{joshi2022efficientgnns}, including lightweight models \cite{sign_icml_grl2020, tailor2021adaptive, li2021gnn1000}, neural architecture search \cite{gao2020graph, zhao2021probabilistic} and quantization techniques \cite{zhao2020learned, tailor2021degreequant}.

\textbf{Knowledge Distillation.}
KD \cite{hinton2015distilling} is a learning paradigm for improving the performance of lightweight `student' models by aligning their outputs to those from cumbersome `teacher' models.
Recent literature has focused on augmenting Hinton et al.'s \cite{hinton2015distilling} logit-based approach with \textit{representational knowledge} from the teacher's latent feature vectors.
Following the taxonomy in \cite{gou2021knowledge}, \textit{feature-based} distillation techniques train the student to imitate the teacher's intermediate feature vectors or attention maps via direct regression \cite{romero2014fitnets, zagoruyko2016paying, li2017mimicking, wang2020defense}.
Due to mismatches in representational capacity between the teacher and student, feature-based distillation may not improve the performance of lightweight students. 
As an alternative, \textit{relation-based} distillation preserves metrics among feature vectors from the teacher's representation space to that of the student, \textit{e.g.} semantic similarity graphs or pairwise distances among samples from the dataset \cite{liu2019knowledge, park2019relational, tung2019similarity}.
The Local Structure Preserving loss (LSP) \cite{yang2020distilling} and its global variant (GSP) extend relation-based distillation for graph data.

Motivated by the success of contrastive objectives for self-supervised learning \cite{oord2018representation, wu2018unsupervised, chen2020simple}, 
Tian et al. \cite{tian2019contrastive} proposed Contrastive Representation Distillation (CRD) for 2D image classification.
In this work, we found that the sample-level CRD objective is not effective for distilling node-level representations built by GNNs. 
We introduce Graph Contrastive Representation Distillation (G-CRD), a node-level contrastive distillation objective tailored to GNNs.


\begin{figure}[h!]
    \centering
    \includegraphics[width=0.6\linewidth]{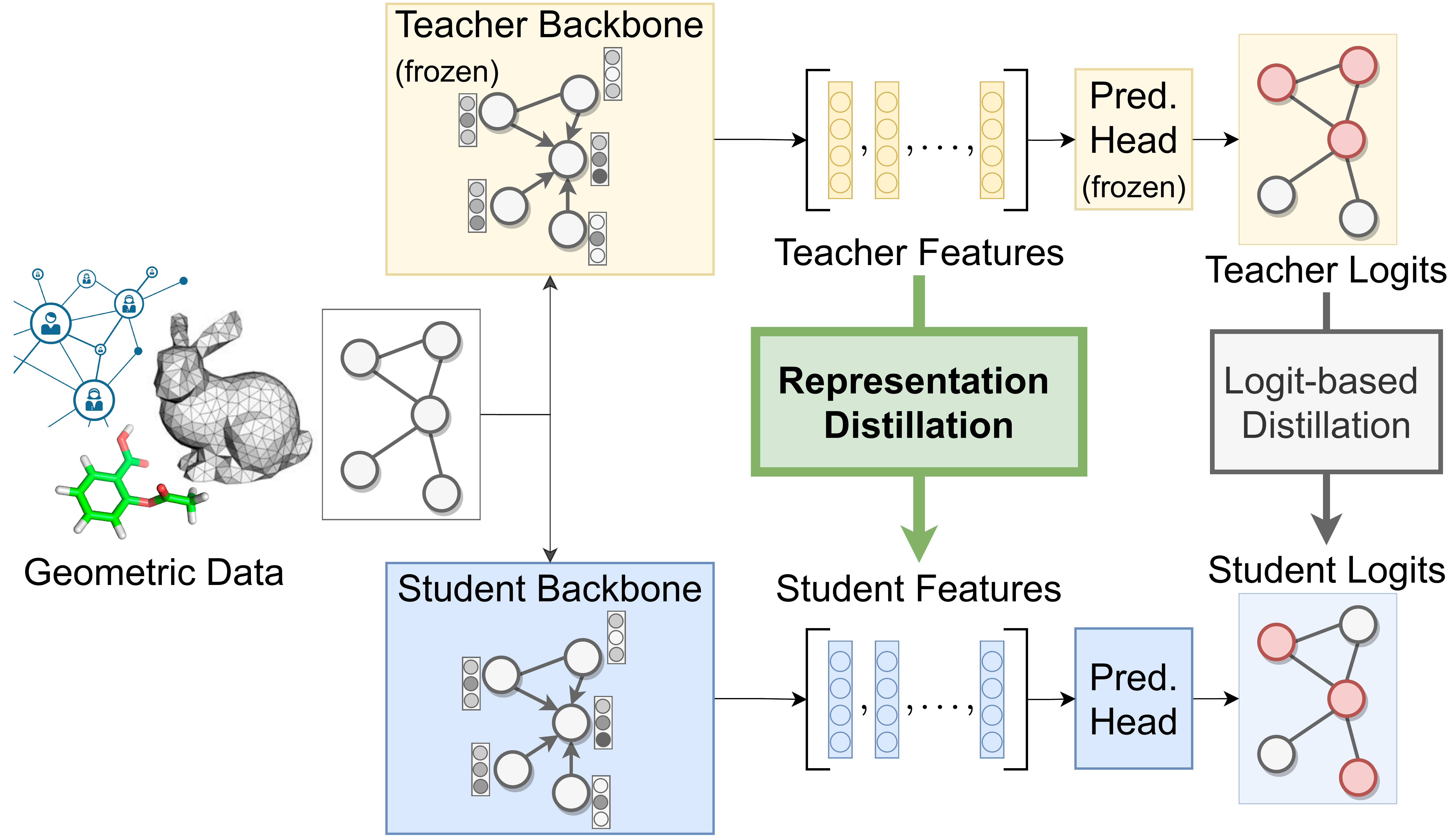}
    \caption{
    \textbf{Knowledge distillation pipeline for graph neural networks.} 
    Logit-based distillation transfers dark knowledge about the data domain via matching the logits of the lightweight student to those of the cumbersome teacher.
    In this work, we investigate \textit{representation distillation}, which transfers representational knowledge using the teacher and student embedding spaces, instead.}
    \label{fig:pipeline}
\end{figure}


\textbf{Knowledge Distillation and GNNs.}
Local Structure Preserving loss (LSP) \cite{yang2020distilling} is the only distillation technique specifically designed for GNNs and message passing style models, but was evaluated on small-scale synthetic datasets.
This work extends the LSP study in two ways:
(1) We introduce G-CRD, a new representation distillation objectives that preserve global topology and consistently outperform LSP;
and (2) We benchmark GNN distillation on large-scale datasets which evaluate for out-of-distribution generalization, as well as across a wider range of teacher-student combinations.

The original logit-based KD \cite{hinton2015distilling} has been applied to GNNs to improve the training of lightweight models such as graph-agnostic MLPs or label propagation \cite{yan2020tinygnn, zhang2020reliable, yang2021extract, zhang2021graphless}.
These works do not introduce new GNN-specific distillation techniques.
Data-free or teacher-free self-distillation \cite{chen2020self,zhang2020iterative,ijcai2021-0320} has also been used to regularize the training of GNNs in the semi-supervised and self-supervised learning paradigm.

Finally, concurrent works on pre-training GNNs for molecular property prediction have also successfully employed contrastive objectives similar to G-CRD to transfer knowledge about 3D molecular geometry into GNNs \cite{stark2021_3dinfomax, liu2021pretraining3d}, but do not focus on model compression.


\section{Teacher and Student Models}
\label{app:models}

Tab.\ref{tab:models} summarizes our choice of teachers and students. 
We benchmark distillation techniques across \textit{heterogeneous} GNN architectures with significant mismatch in terms of depth, parameter count, expressive power, and inference time.
Refer to Sec.\ref{sec:exp:models} for details.

\begin{table}[!ht]
    \centering
    \caption{
    \textbf{Summary of teacher and student model architectures.}
    Peak GPU usage and average inference latency are reported over test set graphs with fixed batch size 1 with single-threading on a single GPU (RTX3090 for ARXIV/MAG, V100 for MOLHIV/S3DIS).
    For MAG, we exclude $136,440,832$ extra embedding dictionary parameters from \#Param. and their lookup time in I.Time.
    }
    \resizebox{\linewidth}{!}{
    \begin{tabular}{clccrcrcrc}
        \toprule
        & \multirow{2}{*}{\textbf{Name}} & \multirow{2}{*}{\textbf{Type}} & \textbf{\#Layer,} & \multicolumn{2}{c}{\multirow{2}{*}{\textbf{\#Parameters}}} & \multicolumn{2}{c}{\textbf{Peak GPU}} & \multicolumn{2}{c}{\textbf{Average}} \\
         & & & \textbf{\#Hidden} & & & \multicolumn{2}{c}{\textbf{Usage}} & \multicolumn{2}{c}{\textbf{I.Time}} \\
        \midrule
        \midrule
        \multirow{4}{*}{\rotatebox[origin=c]{90}{\text{MOLHIV}}} 
         & GIN-E \cite{hu2019strategies} & Teacher & 5, 300 & 3,336,306 & & 1.2GB & & 9ms & \\
         & PNA \cite{corso2020principal} & Teacher & 5, 150 & 2,433,901 & & 1.2GB & & 18ms & \\
         & GCN \cite{kipf2017semi} & Student & 2, 100 & 40,801 & (\green{$\downarrow$98\%}) & 1.1GB & (\green{$\downarrow$9\%}) & 4ms & (\green{$\downarrow$78\%}) \\
         & GIN \cite{xu2018how} & Student & 2, 50 & 10,951 & (\green{$\downarrow$99\%}) & 1.1GB & (\green{$\downarrow$9\%}) & 2ms & (\green{$\downarrow$89\%}) \\
        \midrule
        \multirow{4}{*}{\rotatebox[origin=c]{90}{\text{S3DIS}}}
         & SPVCNN \cite{liu2019point} & Teacher & U-Net & 21,777,421 & & 1.8GB & & 88ms & \\
         & KP-FCNN \cite{thomas2019KPConv} & Teacher & U-Net & 14,082,688 & &  4.8GB & & 227ms & \\
         & PointNet++ \cite{qi2017pointnet++} & Student & U-Net, SSG & 1,400,269 & (\green{$\downarrow$90\%}) & 1.2GB & (\green{$\downarrow$75\%}) & 22ms & (\green{$\downarrow$90\%}) \\
         & MinkNet \cite{choy20194d} & Student & U-Net, 20\%cr. & 852,512 & (\green{$\downarrow$94\%}) & 1.2GB & (\green{$\downarrow$75\%}) & 81ms & (\green{$\downarrow$64\%}) \\
        \midrule
        \multirow{4}{*}{\rotatebox[origin=c]{90}{\text{ARXIV}}}
         & GAT \cite{velickovic2018graph} & Teacher & 3, 750 & 1,441,580 & & 6.8GB & & 93ms & \\
         & GCN \cite{kipf2017semi} & Student & 2, 256 & 43,816 & (\green{$\downarrow$97\%}) & 2.2GB & (\green{$\downarrow$68\%}) & 11ms & (\green{$\downarrow$88\%}) \\
         & GraphSage \cite{hamilton2017inductive} & Student & 2, 256 & 86,824 & (\green{$\downarrow$94\%}) & 2.1GB & (\green{$\downarrow$69\%}) & 5ms & (\green{$\downarrow$95\%}) \\
         & SIGN \cite{sign_icml_grl2020} & Student & 3, 512 & 3,566,128 & (\red{$\uparrow$147\%}) & 9.3GB & (\red{$\uparrow$36\%}) & 5ms & (\green{$\downarrow$95\%}) \\
        \midrule
        \multirow{2}{*}{\rotatebox[origin=c]{90}{\text{MAG}}}
         & R-GCN \cite{schlichtkrull2018modeling} & Teacher & 3, 512 & 5,575,540 & & 20.8GB & & 189ms & \\
         & R-GCN \cite{schlichtkrull2018modeling} & Student & 2, 32 & 169,428 & (\green{$\downarrow$97\%}) & 11.1GB & (\green{$\downarrow$47\%}) & 134ms & (\green{$\downarrow$29\%}) \\
        \bottomrule
    \end{tabular}
    }
    \label{tab:models}
\end{table}


\section{Training Time and GPU Usage}
\label{app:time}

In Tab.\ref{tab:timing}, we report the average training time per epoch as well as GPU usage for distillation techniques.
For the large-scale ARXIV graph, GSP uses significantly higher GPU memory than other approaches due to storing all-to-all pairwise similarity matrices for both teacher and student node embeddings.
For our implementation of LSP (in PyG), the computation of a sparse softmax over local edges in the ARXIV graph is expensive in terms of time cost.

For mini-batched training on molecular graphs from MOLHIV, GSP has a high time cost due to building pairwise similarity matrices for each sample. Although the size of molecular graphs is very small, GSP's GPU memory cost is higher than other approaches, too.

Compared to LSP and GSP, G-CRD is scalable in terms training time and memory usage while consistently improving the performance of student models.
Naturally, G-CRD with the one-layer GCN projection head has a higher time cost than the conventional MLP projection head.
As a reminder, we propose to treat the choice of projection head as a hyperparameter.

\begin{table}[h!]
    \centering
    \caption{
    \textbf{Average training time per epoch and GPU usage for distillation techniques.}
    Time to complete one epoch and GPU usage during training are reported over training set graphs on a single RTX8000 GPU, averaged over 100 epochs.
    }
    \resizebox{0.8\linewidth}{!}{
    \begin{tabular}{llcccc}
        \toprule
        \multicolumn{2}{r}{\textbf{Dataset:}} & \multicolumn{2}{c}{ARXIV} & \multicolumn{2}{c}{MOLHIV} \\
        \multicolumn{2}{c}{\textbf{Teacher} (\#Param):}  & \multicolumn{2}{c}{\textbf{GAT} (1.4M)} & \multicolumn{2}{c}{\textbf{PNA} (2.4M)} \\
        \multicolumn{2}{c}{\textbf{Student} (\#Param):} & \multicolumn{2}{c}{\textbf{GCN} (44K)} & \multicolumn{2}{c}{\textbf{GCN} (40K)} \\
        \multicolumn{2}{c}{ } & Avg. E.Time & Avg. GPU & Avg. E.Time & Avg. GPU \\
        \midrule
        \midrule
        \multirow{2}{*}{\rotatebox[origin=c]{90}{\text{Sup.}}}
        & Teacher & 1.190s & 14.7GB & 42.660s & 2.3GB \\
        & Student & 0.003s & 3.6GB & 10.137s & 2.0GB \\
        \midrule
        \multirow{7}{*}{\rotatebox[origin=c]{90}{\text{Distillation}}}
        & KD \cite{hinton2015distilling} & 0.004s & 4.2GB & 24.086s & 2.1GB \\
        & FitNet \cite{romero2014fitnets} & 0.006s & 5.5GB & 25.535s & 2.1GB \\
        & AT \cite{zagoruyko2016paying} & 0.004s & 5.0GB & 24.499s & 2.1GB \\
        & LSP \cite{yang2020distilling} & 0.100s & 13.1GB & 26.484s & 2.1GB \\
        & GSP & 0.027s & 30.4GB & 45.781s & 2.3GB \\
        & G-CRD (MLP Head) & 0.028s & 6.0GB & 25.464s & 2.1GB \\
        & G-CRD (GCN Head) & 0.066s & 6.4GB & 27.878s & 2.1GB \\
        \bottomrule
    \end{tabular}
    }
    \label{tab:timing}
\end{table} 


\section{Results on PPI}
\label{app:ppi}

In Tab.\ref{tab:ppi}, we attempt to reproduce the results from Yang et. al. \cite{yang2020distilling}, but our codebase\footnote{
We built our implementation upon the official PyG example for PPI.
}
showed significant improvements over their numbers for the same teacher-student pair.
We found negligible performance difference between the two models, even under supervised learning. 
Thus, we redo the experiment with a more reasonable student architecture (a 2 layer GAT instead of 5 layers) and find that G-CRD leads to the best performance, followed by LSP.
We believe small-scale and saturated datasets such as PPI with random splits are not suitable for benchmarking distillation techniques. 

\begin{table}[h!]
    \centering
    \caption{
    \textbf{Node classification on PPI} (metric: F1 (\%)).
    We report the average performance and std. across 10 random seeds.
    }
    \resizebox{0.7\linewidth}{!}{
    \begin{tabular}{llccc}
        \toprule
        \multicolumn{2}{c}{\textbf{Teacher} (\#Layer,\#Param):} & \multicolumn{2}{c}{\textbf{GAT} (3L,3.6M)} & \textbf{GAT} (3L,3.6M) \\
        \multicolumn{2}{c}{\textbf{Student} (\#Layer,\#Param):} & \multicolumn{2}{c}{\textbf{GAT} (5L,160K)} & \textbf{GAT} (2L,180K) \\
        & & Reported & Our Codebase & Our Codebase \\
        \midrule
        \midrule
        \multirow{2}{*}{\rotatebox[origin=c]{90}{\text{Sup.}}}
        & Supervised Teacher & 97.6 & 97.95 \text{$\pm$0.10} & 97.95 \text{$\pm$0.10} \\
        & Supervised Student & 95.7 & 97.90 \text{$\pm$0.45} & 87.55 \text{$\pm$1.33} \\
        \midrule
        \multirow{6}{*}{\rotatebox[origin=c]{90}{\text{Distillation}}}
        & KD \cite{hinton2015distilling} & - & \underline{97.93} \text{$\pm$0.26} & 88.30 \text{$\pm$1.81} \\
        & FitNet \cite{romero2014fitnets} & 95.6 & 97.74 \text{$\pm$0.34} (\red{$\downarrow$}) & 87.46 \text{$\pm$0.78} (\red{$\downarrow$}) \\
        & AT \cite{zagoruyko2016paying} & 95.4 & 97.02 \text{$\pm$0.76} (\red{$\downarrow$}) & 86.82 \text{$\pm$1.27} (\red{$\downarrow$}) \\
        & LSP \cite{yang2020distilling} & \underline{96.1} & 97.81 \text{$\pm$0.31} (\red{$\downarrow$}) & \underline{88.37} \text{$\pm$1.89} (\green{$\uparrow$}) \\
        & GNN-SD \cite{chen2020self} & \textbf{96.2} & - & - \\
        & G-CRD (Ours) & - & \textbf{98.42} \text{$\pm$0.14} (\green{$\uparrow$}) & \textbf{88.38} \text{$\pm$1.36} (\green{$\uparrow$}) \\
        \bottomrule
    \end{tabular}
    }
    \label{tab:ppi}
\end{table}

\section{Ablation Study for G-CRD}
\label{app:gcrd}

This section highlights the key design choices that distinguish G-CRD from Contrastive Representation Distillation (CRD) \cite{tian2019contrastive}, which introduced a contrastive distillation objective for 2D image classification.
CRD is specialized for global per-sample level tasks: it contrasts linearly projected global representations and uses the loss in \eqref{eq:crd}.
On the other hand, G-CRD performs contrastive learning at the node-level using an InfoNCE-style loss \eqref{eq:G-CRD} and a non-linear projection head design \eqref{eq:G-CRD-proj}; it is flexible to handle both node and global-level tasks.

In Tab.\ref{tab:ablation}, we compare G-CRD and our adaptation of CRD to GNNs.
In the first set of rows, we see that contrasting representations at the node-level (as in G-CRD) significantly outperforms global-level contrastive learning (as in CRD) as well as sample-wise node-level (only contrasting among nodes belonging to the same sample).
In the second set of rows, we find that the G-CRD loss outperforms the CRD objective across graph as well as node-level tasks when fixing the contrastive learning at the node-level.
In the third set of rows, we show that the structure-aware GCN projection head boosts performance over a structure-agnostic MLP projection.  
Finally, we show that the full G-CRD implementation consistently outperforms a naive CRD implementation at graph-level tasks.


\begin{table}[t!]
    \centering
    \caption{
    \textbf{Ablation study for G-CRD} (metric: ROC-AUC (\%) for MOLHIV; Accuracy (\%) for ARXIV).
    }
    \resizebox{0.7\linewidth}{!}{
    \begin{tabular}{cccccc}
        \toprule
        \multicolumn{3}{r}{\textbf{Dataset}:} & MOLHIV & MOLHIV & ARXIV \\
        \multirow{2}{*}{\textbf{Repr.}} & \multirow{2}{*}{\textbf{Loss}} & \multirow{2}{*}{\textbf{Proj.}} & \textbf{PNA} (2.4M) & \textbf{PNA} (2.4M) & \textbf{GAT} (1.4M) \\
        & & & \textbf{GCN} (15K) & \textbf{GCN} (40K) & \textbf{GCN} (44K) \\
        \midrule
        \midrule
        \cellcolor{blue!15} Nodes & $\mathcal{L}_{\text{G-CRD}}$ & MLP & \textbf{74.72} \text{$\pm$0.64} & \textbf{75.47} \text{$\pm$0.79} & 71.56 \text{$\pm$0.13} \\
        \cellcolor{blue!15} Nodes (s.w.) & $\mathcal{L}_{\text{G-CRD}}$ & MLP & 74.09 \text{$\pm$1.65} & 74.74 \text{$\pm$0.86} & N.A. \\
        \cellcolor{blue!15} Global & $\mathcal{L}_{\text{G-CRD}}$ & MLP & 73.32 \text{$\pm$1.24} & 73.87 \text{$\pm$1.43} & N.A. \\
        \midrule
        Nodes & \cellcolor{blue!15} $\mathcal{L}_{\text{G-CRD}}$ & MLP & \textbf{74.72} \text{$\pm$0.64} & \textbf{75.47} \text{$\pm$0.79} & \textbf{71.56} \text{$\pm$0.13} \\
        Nodes & \cellcolor{blue!15} $\mathcal{L}_{\text{CRD}}$ & MLP & 73.62 \text{$\pm$1.97} & 75.20 \text{$\pm$1.18} & 71.48 \text{$\pm$0.15} \\
        \midrule
        Nodes & $\mathcal{L}_{\text{G-CRD}}$ & \cellcolor{blue!15} GCN & \textbf{75.11} \text{$\pm$0.73} & \textbf{75.89} \text{$\pm$0.80} & \textbf{71.64} \text{$\pm$0.16} \\
        Nodes & $\mathcal{L}_{\text{G-CRD}}$ & \cellcolor{blue!15} MLP & 74.72 \text{$\pm$0.64} & 75.47 \text{$\pm$0.79} & 71.56 \text{$\pm$0.13} \\
        \midrule
        \multicolumn{3}{c}{\textbf{Full G-CRD}: Nodes + $\mathcal{L}_{\text{G-CRD}}$} + GCN Proj. & \textbf{75.11} \text{$\pm$0.73} & \textbf{75.89} \text{$\pm$0.80} & 71.64 \text{$\pm$0.16} \\
        \multicolumn{3}{c}{\textbf{Full CRD}: Global + $\mathcal{L}_{\text{CRD}}$} + Lin. Proj. & 73.67 \text{$\pm$1.55} & 73.37 \text{$\pm$2.08} & N.A. \\
        \bottomrule
    \end{tabular}
    }
    \label{tab:ablation}
\end{table} 



\section{Ablation Study for GSP}
\label{app:gsp}

In Tab.\ref{tab:gsp-ablation}, we ablate the choice of kernel and metric for GSP, which was proposed as a global structure preserving extension of LSP \cite{yang2020distilling}.
Like LSP, the RBF kernel lead to the best overall performance.
Unlike LSP, we found directly matching teacher and student pairwise similarity matrices via MSE to perform better than normalizing the similarities via softmax followed by KL-divergence.
We attribute this to the sparsity of the signal in the resulting matrices after the softmax.

\begin{table}[t!]
    \centering
    \caption{
    \textbf{Ablation study for GSP} (metric: ROC-AUC (\%) for MOLHIV; Accuracy (\%) for ARXIV).
    }
    \resizebox{0.65\linewidth}{!}{
    \begin{tabular}{ccccc}
        \toprule
        \multicolumn{2}{r}{\textbf{Dataset}:} & MOLHIV & MOLHIV & ARXIV \\
        \multirow{2}{*}{\textbf{Kernel}} & \multirow{2}{*}{\textbf{Metric}} & \textbf{GIN-E} (3.3M) & \textbf{GIN-E} (3.3M) & \textbf{GAT} (1.4M) \\
        & & \textbf{GCN} (15K) & \textbf{GCN} (40K) & \textbf{GraphSage} (87K) \\
        \midrule
        \midrule
        \cellcolor{blue!15} Euclidean & MSE & 62.73 \text{$\pm$3.30} & 68.67 \text{$\pm$2.42} & 70.89 \text{$\pm$0.22} \\
        \cellcolor{blue!15} Linear & MSE & \textbf{73.01} \text{$\pm$1.00} & 74.90 \text{$\pm$1.60} & 70.98 \text{$\pm$0.33} \\
        \cellcolor{blue!15} Polynomial & MSE & 72.58 \text{$\pm$0.82} & 74.07 \text{$\pm$1.60} & 71.00 \text{$\pm$0.18} \\
        \cellcolor{blue!15} RBF & MSE & 72.83 \text{$\pm$1.30} & \textbf{75.12} \text{$\pm$1.27} & \textbf{71.05} \text{$\pm$0.20} \\
        \midrule
        RBF & \cellcolor{blue!15} KL-div. & 72.46 \text{$\pm$0.81} & 74.33 \text{$\pm$2.24} & 70.96 \text{$\pm$0.08} \\
        RBF & \cellcolor{blue!15} MSE & \textbf{72.83} \text{$\pm$1.30} & \textbf{75.12} \text{$\pm$1.27} & \textbf{71.05} \text{$\pm$0.20} \\
        \bottomrule
    \end{tabular}
    }
    \label{tab:gsp-ablation}
\end{table} 


\newpage
\section{Pseudocode}

Listings 1 and 2, provide PyTorch-style pseudocode for the GSP and G-CRD objectives, respectively.
Our code is available at \url{https://github.com/chaitjo/efficient-gnns}

\begin{lstlisting}[language=Python, caption=PyTorch-style pseudocode for the Global Structure Preserving objective with cosine pairwise similarity kernel.]
import torch
import torch.nn.functional as F

def gsp_cosine_criterion(
    student_feat, # features from student model
    teacher_feat, # features from teacher model
):
    
    # L2-normalize student features, dim: (n, d_s)
    student_feat = F.normalize(
        student_feat, p=2, dim=-1
    )
    
    # L2-normalize teacher features, dim: (n, d_t)
    teacher_feat = F.normalize(
        teacher_feat, p=2, dim=-1
    )
    
    # all pairwise cosine similarities for student
    # dim: (n, d_s).(d_s, n) -> (n, n)
    student_pw_sim = torch.mm(
        student_feat, student_feat.transpose(0, 1)
    )
    
    # all pairwise cosine similarities for teacher
    # dim: (n, d_t).(d_t, n) -> (n, n)
    teacher_pw_sim = torch.mm(
        teacher_feat, teacher_feat.transpose(0, 1)
    )
    
    # regression to match pairwise similarities
    loss_gsp = F.mse_loss(
        student_pw_sim, teacher_pw_sim
    )
    
    return loss_gsp
\end{lstlisting}

\newpage

\begin{lstlisting}[language=Python, caption=PyTorch-style pseudocode for the Graph Contrastive Representation Distillation objective.]
import torch
import torch.nn.functional as F

def gcrd_criterion(
    student_feat,      # features from student model
    teacher_feat,      # features from teacher model
    student_proj_head, # learnt projection head
    teacher_proj_head, # learnt projection head
    nce_tau            # InfoNCE temperature
):
    
    # project and L2-normalize student features
    # dim: (n, d_s) -> (n, d)
    student_feat = F.normalize(
        student_proj_head(student_feat), 
        p=2, dim=-1
    )
    
    # project and L2-normalize teacher features
    # dim: (n, d_t) -> (n, d)
    teacher_feat = F.normalize(
        teacher_proj_head(teacher_feat), 
        p=2, dim=-1
    )
    
    # cosine similarities b/w student and teacher
    # dim: (n, d).(d, n) -> (n, n)
    nce_logits = torch.mm(
        student_feat, teacher_feat.transpose(0, 1)
    )
    
    # prepare labels for correspondence task
    # dim: (n, 1)
    nce_labels = torch.arange(student_feat.shape[0])
    
    # temperature-scaled softmax cross entropy
    loss_nce = F.cross_entropy(
        nce_logits/ nce_tau, nce_labels
    )
    
    return loss_nce
\end{lstlisting}


\end{document}